\documentclass{article}

\usepackage[nonatbib,preprint]{neurips_2021}
\usepackage[utf8]{inputenc} %
\usepackage[T1]{fontenc}    %
\usepackage{hyperref}
\usepackage{url}            %
\usepackage{booktabs}       %
\usepackage{amsfonts}       %
\usepackage{amsmath}
\usepackage{amssymb}
\usepackage{amsthm}
\usepackage{nicefrac}       %
\usepackage{microtype}      %
\usepackage[table]{xcolor}         %
\usepackage{algorithm}
\usepackage{algorithmic}
\usepackage[utf8]{inputenc} %
\usepackage[T1]{fontenc}    %
\usepackage{microtype}
\usepackage{graphicx}
\usepackage{subfigure}
\usepackage{booktabs} %
\usepackage[
    maxbibnames=9, maxcitenames=1,
    natbib,
    citestyle=numeric-comp,
    sorting=none,
    sortcites=true
]{biblatex}
\usepackage{hyperref}       %
\usepackage{url}            %
\usepackage{microtype}      %

\usepackage{bm}
\usepackage{pifont}
\usepackage{mathtools}
\usepackage{comment}
\usepackage{booktabs}
\usepackage{tikz}
\usepackage{physics}
\usepackage{mathdots}
\usepackage{yhmath}
\usepackage{cancel}
\usepackage{siunitx}
\usepackage{array}
\usepackage{multirow}
\usepackage{gensymb}
\usepackage{tabularx}
\usepackage{wrapfig}
\usetikzlibrary{fadings}
\usetikzlibrary{patterns}
\usetikzlibrary{shadows.blur}
\usetikzlibrary{shapes}
\usepackage{color}
\usepackage{adjustbox}
\usepackage{threeparttable}   %
\usepackage{acronym} %
\usepackage[page]{appendix}
\usepackage{calc}

\newtheorem*{theorem*}{Theorem}
\newtheorem*{definition*}{Definition}

\newcommand\Bh{\bm{h}}

\newcommand\Bm{\bm{m}}

\newcommand\Bp{\bm{p}}
\newcommand\Bq{\bm{q}}
\newcommand\Br{\bm{r}}

\newcommand\Bt{\bm{t}}

\newcommand\Bv{\bm{v}}
\newcommand\Bw{\bm{w}}
\newcommand\Bx{\bm{x}}
\newcommand\By{\bm{y}}

\newcommand\BL{\bm{L}}

\newcommand\BT{\bm{T}}

\newcommand\BW{\bm{W}}
\newcommand\BX{\bm{X}}

\newcommand\Bxi{\bm{\xi}}

\newcommand\BOn{\bm{1}}

 \newcommand{\dR}{\mathbb{R}}

\newcommand{\rE}{\mathrm{E}}

\newcommand\nn{\mathrm{new}}

\newcommand{\soft}{\mathrm{softmax}}

\definecolor{lightblue}{RGB}{32, 194, 217}
\definecolor{jku_red}{RGB}{217, 92, 76}
\definecolor{jku_blue}{RGB}{0, 132, 187}
\definecolor{jku_green}{RGB}{91, 167, 85}
\definecolor{jku_yellow}{RGB}{241, 188, 63}
\definecolor{jku_cyan}{RGB}{79,176,191}
\definecolor{jku_gray}{RGB}{125,130,140}
\definecolor{jku_lightgreen}{RGB}{191,206,82}
\definecolor{jku_violett}{RGB}{174,97,157}

\newcommand{\janssen}{High Dimensional Biology and Discovery Data Sciences \\ Janssen Research \& Development, Janssen Pharmaceutica N.V.}
\newcommand{\microsoft}{Microsoft Research}

\title{Modern Hopfield Networks for Few- and Zero-Shot Reaction 
Template Prediction}
\author{\vspace{0.1cm}
    Philipp Seidl\footnotemark[1] \quad
    Philipp Renz\footnotemark[1] \quad \\
    \vspace{0.1cm} \bf
    Natalia Dyubankova \footnotemark[3]\quad
    Paulo Neves\footnotemark[3] \quad
    Jonaes Verhoeven\footnotemark[3] \quad 
    Marwin Segler\footnotemark[4] \quad 
    Jörg K. Wegner\footnotemark[3] \quad 
    \\ \vspace{0.1cm} \bf
    Sepp Hochreiter\footnotemark[1]~$~^{,}$\footnotemark[2]
    Günter Klambauer\footnotemark[1]  \quad \\
  \footnotemark[1]~~ELLIS Unit Linz, LIT AI Lab, Institute for Machine Learning,\\
                  ~~Johannes Kepler University Linz, Austria\\
  \footnotemark[2]~~Institute of Advanced Research in 
                    Artificial Intelligence (IARAI) \\
  \footnotemark[3]~~\janssen \\
  \footnotemark[4]~~\microsoft              
}

\begin{document}
\maketitle
\begin{abstract}
Finding synthesis routes for molecules of
interest is an essential step in the discovery
of new drugs and materials.
To find such routes, 
computer-assisted synthesis planning (CASP)
methods are employed, which rely on a model 
of chemical reactivity.
In this study, we model single-step
retrosynthesis in a template-based approach
using modern Hopfield networks (MHNs).
We adapt MHNs to associate different modalities, reaction templates and molecules, 
which allows the model to leverage structural information about reaction templates.
This approach significantly improves the performance of template relevance prediction, 
especially for templates with few or zero training examples.
With inference speed several times faster than baseline 
methods, we improve predictive performance for top-$\mathrm{k}$ 
exact match accuracy for
$\mathrm{k}\geq5$ in the retrosynthesis benchmark USPTO-50k.
Code to reproduce the results will be available at \href{https://github.com/ml-jku/mhn-react}{github.com/ml-jku/mhn-react}.
\end{abstract}

\section{Introduction} 
\begin{wrapfigure}[19]{r}{0.5\textwidth}
  \vspace{-0.6cm}
  \begin{center}
  \includegraphics[width=0.48\textwidth]{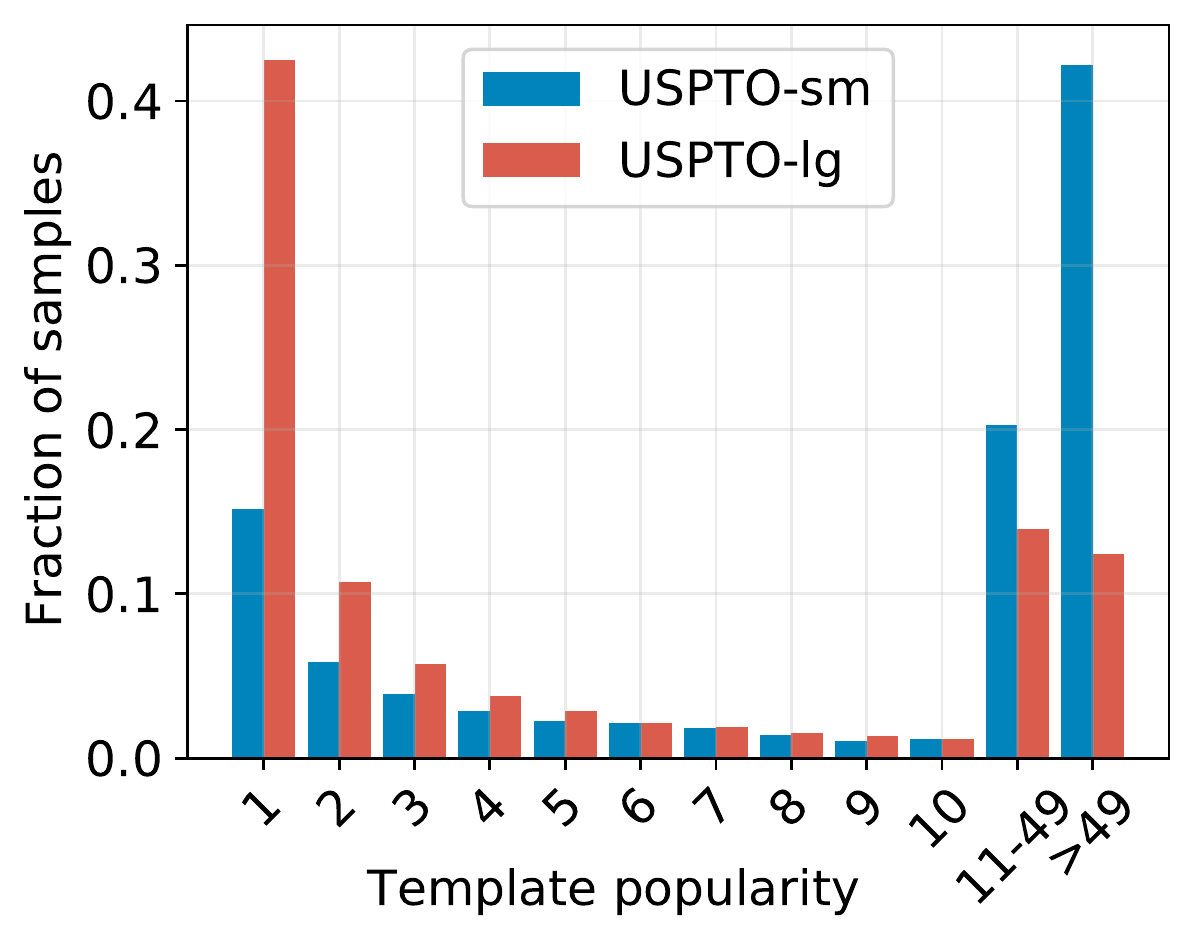}
  \caption{Histogram showing the fraction of samples for different template frequencies. The leftmost red bar indicates that over 40\% of chemical reactions of USPTO-lg have a unique reaction template. The majority of reaction templates are rare.
  \label{fig:template_dist}}
  \end{center}
  \vspace{-0.5cm}
\end{wrapfigure}
The design of a new molecule starts with
an initial idea of a chemical structure with hypothesized
desired properties \citep{lombardino2004role}. 
Desired properties might be the inhibition of a disease or a virus
in drug discovery or thermal stability in material science 
\citep{lu2020interpretable,mayr2018large}.
From the design idea of the molecule,
a virtual molecule is constructed, the properties of which can then
be predicted by means of computational methods \citep{mccammon1987computeraided,ng2015challenges}. 
However, to test its properties and to finally make use of
it, the molecule has to be made physically available through
chemical synthesis. 
Finding a synthesis route for a given molecule is a
multi-step process that is considered highly complex \citep{corey1969computerassisted}. 

To aid in finding synthesis routes, chemists have
resorted to computer-assisted synthesis planning (CASP)
methods \citep{corey1969computerassisted,szymkuc2016computerassisted}.
Chemical synthesis planning is often viewed in the 
retrosynthesis setting in which a molecule of interest is recursively 
decomposed into less complex molecules
until only readily available precursor molecules remain
\citep{segler2018planning}.
Such an approach relies on a single-step retrosynthesis model, which,
given a product, tries to propose sets of reactants from which it can
be synthesized. 
Early methods modelled chemical reactivity using rule‐based expert
systems \citep{szymkuc2016computerassisted}.
These methods, however, require extensive manual curation
\citep{struble2020current, segler2017neuralsymbolic, segler2018planning}.
Recently there have been increased efforts to
model chemical reactivity from reaction databases
using machine learning methods \citep{coley2018machine,segler2018planning, coley2017computerassisted, schwaller2019molecular}.

These efforts to model chemical reactions 
encompass a variety of different approaches.
In one line of methods
\citep{nam2016linking,schwaller2018found, liu2017retrosynthetic, schwaller2019molecular,karpov2019transformer, tetko2020stateoftheart},
the SMILES representation \citep{weininger1988smiles} of the reactants given that of the product is predicted, using architectures proposed initially for the translation between natural languages \citep{vaswani2017attention,sutskever2014sequence}.
Others exploit the graph structure of molecules and model the task using graph neural networks \citep{shi2020graph, somnath2020learning}.
A prominent line of work makes use of \emph{reaction templates} which are graph transformation rules that encode connectivity changes between atoms during a chemical reaction.

In a template-based approach, reaction templates are first extracted from 
a reaction database or hand-coded by a chemist.
If the product side of a template is a sub-graph of a molecule, the template is
called applicable to the molecule, and can be used to transform it into a 
reactant set. However, even if a template can be applied to a molecule, the resulting reaction might not be viable in the lab
\citep{segler2017neuralsymbolic}. 
Hence a core task, which we refer to as template-relevance prediction, in such an approach
is to predict with which templates a molecule can be combined with to yield a viable reaction.
In prior work, this problem has often been tackled using a machine learning
methods that are trained at this task on a set of recorded reactions
\citep{fortunato2020data,segler2017neuralsymbolic,segler2018planning, wei2016neural,segler2018planning,bjerrum2020artificial,dai2020retrosynthesis, baylon2019enhancing, coley2017retrosim, ishida2019prediction}. 

\begin{figure}
    \hspace{-0.6cm}
    \resizebox{\textwidth}{!}{%
        \input{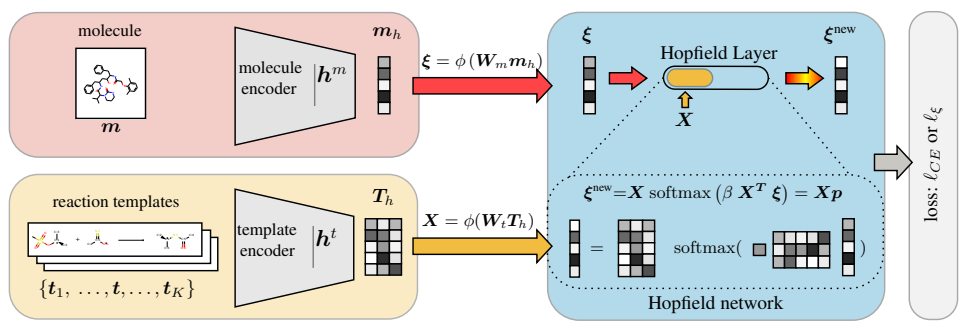}
    }
    \caption{Simplified depiction of our approach. Standard approaches only encode the molecule and predict a fixed set of templates. In our MHN-based approach, the templates are also encoded and transformed to stored patterns via the template encoder. The Hopfield layer learns to associate
    the encoded input molecule, the state pattern $\boldsymbol{\xi}$,
    with the memory of encoded templates, the stored patterns $\boldsymbol{X}$. Multiple Hopfield layers can operate in parallel or be stacked using different encoders. 
    }
    \label{fig:overview}
\end{figure}

Template-based methods %
usually view the problem as a classification task in which the templates are modelled as
distinct categories. However, this can be problematic as 
automatic template extraction leads to many templates
that are represented by few training
samples \citep{fortunato2020data, segler2018planning}.
\citet{somnath2020learning} argue that template-based approaches suffer from bad performance, particularly for rare reaction templates.
\citet{struble2020current} note that ML has not been applied successfully for CASP in low-data regimes. 
To address the low-data issue, \citet{fortunato2020data}
pretrained their template-relevance model to predict which templates are
applicable and then fine-tuned it on recorded reactions in a database. This
improved template-relevance prediction, especially for rare templates, as well as
the average applicability of the top-ranked templates.
Overall, a challenge of template-based methods arises from modelling reaction templates as 
distinct categories,
which leads to many classes with few examples
(see also Section~\ref{sec:motivation}).

To avoid the above-mentioned problems, 
we propose a new model that does not consider templates as distinct
categories, but can leverage structural information about the template.
This allows for generalization over templates and improves performance in the
tasks defined in \citep{fortunato2020data}, especially for templates with few
training samples and even for unseen templates.
This model learns to associate relevant templates to product molecules 
using a Modern Hopfield Network
 \citep{ramsauer2020hopfield,widrich2020modern}.
To this end, we adapted MHNs to associate objects of different modalities, namely input molecules, and reaction templates. 
In contrast to popular ML approaches, in which variable
or input-dependent subsets of the data 
are associated \citep{vaswani2017attention, ramsauer2020hopfield, chen2020simple,radford2021learning}, our 
architecture maintains a fixed set 
of representations, considered as a 
static memory independent of the input.

In this study, we propose a template-based method, which 
are often reported to be computationally
expensive because of the NP-complete subgraph-isomorphism calculations involved
in template execution \citep{shi2020graph,somnath2020learning,fortunato2020data,bjerrum2020artificial}.
To address this issue \citet{bjerrum2020artificial,fortunato2020data} trained
neural networks to predict which templates are applicable given a molecule to
filter inapplicable templates during inference. 
We find that using a substructure
screen, i.e. a fast check of a necessary condition
for a graph to be a subgraph of another,
improves inference speed which may also be of
interest for other template-based methods.

\textbf{Contributions.} 
In Section~\ref{sec:mhn} we adapt modern Hopfield networks to 
associate different data modalities and to use a static memory 
resulting in a 
novel template-based approach to chemical synthesis. 
In Section~\ref{sec:experiments} we demonstrate that our architecture 
improves predictive performance for template relevance prediction and 
single-step retrosynthesis.
In Section~\ref{sec:speed} we investigate inference speed and show that our method is several times faster than baseline methods. 

\section{Background: single-step
retrosynthesis\label{sec:background}} 
The goal of \emph{single-step retrosynthesis} is to predict sets of molecules that
react to a given product. As a molecule can be synthesized in various ways,
this represents a one-to-many task. A model is thus expected to output
a list of reactant sets sorted by a belief that a solution is correct. 
Performance in this setting is usually measured by \emph{reactant top-$\mathrm{k}$ accuracy}
using a reaction database. 
This metric measures the fraction of samples for which, given the product of
a recorded reaction, the recorded reactants are among the top-$\mathrm{k}$ predictions.
Given the one-to-many setting small values of $\mathrm{k}$ might not be an optimal
choice as there might exist scenarios where a good model receives low scores.
Choosing $\mathrm{k}$ too large might result in a metric that is too easy to optimize.

Template-based approaches predict reactant sets via reaction templates. 
A reaction template encodes atom connectivity changes during a chemical
reaction and can be used to transform a product molecule to reactants,
$ \Bm \ \xrightarrow[]{\ \ \Bt \ \ } \ \Br $,
where $\Bm$ is a product molecule, $\Br$ represents a set of reactants
and $\Bt$ a reaction template.
The product side of a template encodes at which position
in a molecule the template can be applied. A necessary 
condition for this is that the product side of the template is a substructure
of the molecule of interest. If this is the case a template is said to be
\emph{applicable} to the molecule.
The product sub-graph is then transformed according to the reactant side of the
template and an atom-mapping between the two sides. Templates can be either hand-coded or automatically extracted from
reaction databases, which yields an ordered set of $K$ unique templates 
$\BT=\{\Bt^k\}_{k=1}^K$.

The aim of \emph{template-relevance prediction} is to predict which templates 
result in a feasible reaction given a product. If this is the case we say that a template is \emph{relevant} to a molecule.
While applicability is a necessary condition for relevance, it 
ignores the context of the whole molecule and thus
substructures that might conflict with the encoded reaction
\citep[Fig. 1]{segler2017neuralsymbolic}.
In practice applicability gives poor performance at
relevance prediction (see Table~\ref{tab:accuracy}).
To evaluate template-relevance predictions we use 
\emph{template top-$\mathrm{k}$ accuracy}, which given the product of a recorded reaction 
measures the fraction of samples for 
which the template extracted from the recorded reaction is among the top-$\mathrm{k}$
predicted ones. 

Given relevance predictions for a product, reactant sets are obtained by 
executing top-scoring templates. 
We do not permit relevance prediction to rely on 
applicability calculations because it is relatively slow to compute. Via this constraint, 
template top-$\mathrm{k}$ accuracy also incorporates information about the models ability 
to filter out non-applicable templates. 
This information might be lost in reactant accuracy as template execution relies
on a check for applicability.
Other differences between the reactant/template accuracy can arise from multiple
locations in which the correct template may be applied or incorrect templates
leading to the correct reactants.

\section{Modern Hopfield networks for reaction template prediction\label{sec:mhn}}
\label{sec:motivation}
\textbf{Motivation of our approach.}
Many template-based methods \citep{fortunato2020data, segler2017neuralsymbolic, segler2018planning,wei2016neural,bjerrum2020artificial} predict templates using 
\begin{align}
\label{eq:simple_model}
\hat \By &= \soft(\BW \  \Bh^m (\Bm)),  %
\end{align}
where $\Bh^m(\Bm)$ is a neural network that maps a molecule
representation
to a vector of size $d_m$, which we call \emph{molecule encoder}.
Multiplication with $\BW \in \dR^{K\times d_m}$ yields a score for each
template $\Bt_1,\ldots, \Bt_K$.
These scores are then normalized using the softmax function, which yields the vector $\hat \By \in \dR^K$.
In this setting, different templates are viewed as distinct categories or classes,
which makes the
model ignorant of similarities between classes which prevents generalization
over templates. 
The high fraction of samples in reaction
data sets that have a unique template can be problematic
because they cannot contribute to
performance. This problem might also appear for
templates occurring only a few times, but to a lesser
extent.

Instead of learning the rows of $\BW$ independently, 
one could map each
template to a vector of size $d_t$ 
using a \emph{template encoder}, $\Bh^t$,
and concatenate them row-wise to obtain
$\BT_{h} = \Bh^t(\BT) \in \dR^{K \times d_t}$.
If $d_m=d_t$, replacing $\BW$ in the equation above yields
\begin{align}
 \label{eq:motiv}
  \hat \By \ &= \ \soft(\BT_{h} \ \Bh^m (\Bm)),
\end{align}
which associates the molecule $\Bm$ with each template via the dot
product of their representations. 
This allows generalization across templates because 
the structure of the template is used to represent 
the class and the model can leverage similarities between templates.
We adapt 
modern Hopfield networks \citep{ramsauer2020hopfield,ramsauer2021hopfield} 
to generalize this association of the two modalities,
molecules and reaction templates.

A component of the model proposed in \citet{dai2020retrosynthesis} 
is similar to our approach as it also makes use of the templates' structures. 
Our approach also resembles the approach taken in CLIP \citep{radford2021learning}
and ConVIRT \citep{zhang2020contrastive}
in which associated pairs of images and texts are
contrasted against non-associated pairs. 
Our adaption of MHNs to maintain a static memory
complements previous 
contrastive learning \citep{hadsell2006dimensionality,chen2020simple} 
approaches using a memory \citep{wu2018unsupervised,misra2020selfsupervised,he2020momentum,tian2020contrastive}.

\textbf{Modern Hopfield networks.}
Recently \citet{ramsauer2020hopfield,ramsauer2021hopfield} introduced
modern Hopfield Networks (MHN) which 
are well-suited for learning associations between sets
\citep{widrich2020modern}.
We use an MHN approach to generalize Eq.~\eqref{eq:motiv} 
to a more expressive association mapping
\begin{align}
\label{eq:model}
\hat \By \ &= \ g(\Bh^m(\Bm),\Bh^t(\BT)), 
\end{align}
where $g$ consists of Hopfield layers based on MHNs.
An MHN is an associative memory that 
stores patterns $\BX \in \dR^{d \times K}$ \ \footnote{Note that $K$ in this work corresponds  to $N$ in \citep{ramsauer2020hopfield}.} 
and iteratively updates a state pattern according to:
\begin{align}
\label{eq:update}
\Bxi^{\nn} \ &= \ \BX \Bp \ = \   
\BX \soft ( \beta \BX^T \Bxi) \ , %
\end{align}
where $\Bp$ is called the vector of \emph{associations} and
$\beta>0 $ is a scaling parameter (inverse temperature).
The update rule converges globally 
to stationary points of the energy function,
$\rE \ = \ - \ \mathrm{lse}(\beta ,\BX^T \Bxi) \ + \
  \frac{1}{2} \Bxi^T \Bxi  + C$, where 
$\mathrm{lse}$ is the log-sum-exponential function, 
and $C$ is a constant \citep[Theorem 1 and 2]{ramsauer2020hopfield}. 
To adapt MHNs to reaction template prediction, 
the Hopfield layer learns both a 
molecule representation used for the state pattern $\Bxi \in \dR^d$ and 
reaction template representations used for the stored patterns $\BX$.

\textbf{Model architecture with parallel or stacked Hopfield layers.} 
Our model architecture consists of three main
parts: a) a molecule encoder, b) a reaction template encoder, and
c) one or more stacked or parallel Hopfield layers. 
Firstly, we use a molecule encoder function that 
learns a relevant representation
for the task at hand. For this we use a
fingerprint-based, e.g.\ ECFP \citep{rogers2010extendedconnectivity}, 
fully-connected NN, $\Bh^m_{\Bw}(\Bm)$ with weights $\Bw$. 
The molecule encoder maps a molecule to its 
representation $\Bm_h=\Bh^m_{\Bw}(\Bm)$ of dimension $d_m$.

Secondly, we use the reaction template encoder
$\Bh^t_{\Bv}$ with parameters $\Bv$ to learn relevant representations of templates.
Here, we also use a fully-connected NN with \emph{template fingerprints} as input. These fingerprints are described in Appendix~Section~\ref{app:experiments}.
This function is applied
to all templates $\BT$ and the resulting vectors are 
concatenated column-wise into a
matrix $\BT_h = \Bh^t_{\Bv}(\BT)$ with
shape $(d_t, K)$.

Finally, we use a single or several stacked or parallel
Hopfield layers $g(.,.)$ 
to associate a molecule with all templates in the memory.
The Hopfield layer consists of fully-connected layers 
with adaptive weight matrices $\BW_m, \BW_t$
to map both molecule
representations and template representations
to a $d$-dimensional space: 
$\Bxi = \phi(\BW_m \ \Bh^m_{\Bw} (\Bm))$ and
$\BX = \phi(\BW_t \ \Bh^t_{\Bv}(\BT))$, where $\BW_m$ 
and $\BW_t$ are adaptive weight matrices with $d$ rows, 
and $\phi$ is an activation function. 
Hopfield layers comprise layer normalization \citep{ba2016layer} for $\Bxi$ and $\BX$
which is included as a hyperparameter. 
We also consider the scaling parameter $\beta$ as a hyperparameter.
The Hopfield layer then employs the update rule Eq.~\eqref{eq:update}
through which the updated representation 
of the product molecule $\Bxi^{\mathrm{new}}$ and 
the vector of associations $\Bp$ is obtained. 
If multiple Hopfield layers are stacked, $\Bxi^{\nn}$ enters
the next Hopfield layer, for which additional template encoders supply the template
representations. 
Parallel Hopfield layers use the same
template encoder, but learn different projections $\BW_t,\BW_m$, 
which is analogous to the heads in Transformer networks. 

The simple model (Eq.~\ref{eq:motiv})
is a special case of our MHN and recovered 
if 
(a) $\BW_t$ and $\BW_m$ are the identity matrices 
and $d_t=d_m=d$,
(b) the Hopfield network is constrained to a single update,
(c) Hopfield networks are not stacked, i.e. there is only a 
single Hopfield layer
(d) the scaling parameter $\beta=1$,
(e) layernorm learns zero mean and unit variance and does
not use its adaptive parameters, and 
(f) the activation function $\phi$ is the identity.
The standard DNN model (Eq.~\ref{eq:simple_model}) is 
recovered if additionally the reaction templates are one-hot 
encoded, and the template encoder is linear. 

In this study, we tested fingerprint-based fully-connected networks for 
the molecule and template encoder.
In principle, one could use any mapping from 
molecules/templates to vector-valued representations for these components, for example raw fingerprints, 
graph neural networks \citep{gilmer2017neural} or 
SMILES/SMARTS-based RNNs \citep{mayr2018largescale} or Transformers \citep{vaswani2017attention}.

\textbf{Loss function and optimization.}
\label{sec:loss}
We provide a general definition of the loss in terms 
of retrieved patterns and details in Appendix~Section~\ref{app:loss}. 
Here we supply a simple view of the case in which 
a single correct reaction template is assigned to each molecule.
Given a training pair $(\Bm,\Bt)$ and the set of all
templates $\BT$, the model should assign high probability 
to $\Bt$ based on $\Bm$ and $\BT$.
We encode this objective by the negative log-likelihood: $-\log p(\Bt \mid \Bm, \BT)$. 
The probability of each template in $\BT$ is encoded by
the corresponding element of the vector of associations $\Bp$  of the last Hopfield layer.
In case of multiple parallel Hopfield layers, we use 
average pooling across the vectors $\Bp$ supplied from 
each layer. 

\begin{wrapfigure}[17]{r}{0.55\textwidth}
    \vspace{-0.65cm}
    \begin{minipage}{0.55\textwidth}
      \begin{algorithm}[H]
        \caption{MHNs for template prediction (simplified)\label{alg:mhn}}
          \begin{algorithmic}
          \STATE  {\color{lightblue} \# \texttt{mol\_encoder()}} Maps to dimension $d_m$.
          \STATE {\color{lightblue} \# \texttt{template\_encoder()}} Maps to dimension $d_t$.
          \STATE {\color{lightblue} \# \texttt{m\_train, t\_train}} --- pair of product molecule and reaction template from training set
          \STATE {\color{lightblue} \# \texttt{T}} --- set of $K$ reaction templates including \texttt{t\_train}
          \STATE {\color{lightblue} \# \texttt{d}} --- dimension of Hopfield space
          \STATE {\color{lightblue} \texttt{\#\# forward pass}}
          \STATE \texttt{T\_h = template\_encoder(T){\color{lightblue} \#[d\_t,K]}} 
          \STATE \texttt{m\_h = mol\_encoder(m\_train){\color{lightblue} \#[d\_m,1]}} 
          \STATE \texttt{xinew,p,X = Hopfield(m\_h,T\_h,dim=d)} 
          \STATE \texttt{p = pool(p,axis=1) {\color{lightblue} \#[K]}} 
          \STATE {\color{lightblue} \texttt{\#\# association loss}}
          \STATE \texttt{label = where(T==t\_train){\color{lightblue} \#[K]}} 
          \STATE \texttt{loss = cross\_entropy(p,label)} 
        \end{algorithmic}
      \end{algorithm}
    \end{minipage}
    
  \end{wrapfigure}
The parameters of the model are adjusted 
on a training set using stochastic 
gradient descent on the loss w.r.t.\ $\BW_t,\BW_m, \Bw,\Bv$ 
via the AdamW optimizer \citep{loshchilov2018fixing}. 
We train our model for a maximum of 100 epochs and then 
select the best model with respect to the minimum cross-entropy loss
in the case of template-relevance prediction 
or maximum top-$1$ accuracy for single-step retrosynthesis on the validation set. 
Pseudo-code of the forward-pass 
of our model is presented in Alg.~\ref{alg:mhn}.

\textbf{Regularization.} We use dropout regularization 
in the molecule encoder $\Bh^m$, for the 
template encoder $\Bh^t$ as well as for the representations in the Hopfield layers.
We employ L2 regularization on the parameters.
A detailed list of considered and selected 
hyperparameters is given in Appendix~Tables~\ref{tab:hparams_tr} and \ref{tab:hparams_ssr}.

\textbf{Fingerprint filter (FPF).}
We added a computationally cheap fingerprint-based
substructure screen
as a post-processing step that can filter out a
part of the non-applicable templates.
For each product and the product side of each template,
we calculated a bit-vector using the 
"PatternFingerprint" function from rdkit
\citep{landrum2006rdkit}. Each bit set in this vector
specifies the presence of a substructure.
For a template to be applicable every 
bit set in the template fingerprint also has to be set 
in the product fingerprint, which is 
a necessary condition for subgraphs to match. 
We chose a fingerprint size of 4096 as we did not observe significant 
performance gains for larger sizes, as can be seen in Appendix~Fig.~\ref{fig:result_fp_size}.

\section{Experiments\label{sec:experiments}}
\textbf{Data and preprocessing.} All data sets used in this study are derived 
from the USPTO data set, extracted from the U.S. patent literature by \citet{lowe2012extraction}. 
This data set contains ~1,8M text-mined reaction equations in 
SMILES notation \citep{weininger1988smiles} and consists of reactions recorded in the years from
1976 to 2016. Reaction conditions and process actions are not included.
For evaluating template relevance prediction, we use the preprocessing procedure described in 
\citep{fortunato2020data}. Templates are extracted using rdchiral \citep{coley2019rdchiral}.
This results in two data sets, \emph{USPTO-sm} which is based on USPTO-50k
\citep{schneider2016what} and \emph{USPTO-lg} which is based on USPTO-410k 
\citep{w.coley2019graphconvolutional}.
For evaluating single-step retrosynthesis we use 
USPTO-50k as preprocessed in \citep{coley2017retrosim}.
For this set we also extract templates using rdchiral \citep{coley2019rdchiral},
but only for the train- and validation split to prevent test data leakage.
A detailed description of the data sets and their preprocessing can
be found in Appendix~\ref{app:experiments}.

{\textbf{Compute time and resources.} All experiments were run on different servers with diverse Nvidia GPUs (Titan V 12GB, P40 24 GB, V100 16GB, A100 20GB MIG) 
using  PyTorch 1.6.0 \citep{paszke2019pytorch}. We estimate the total run-time 
to be around 1000 GPU hours. %
A single MHN model can be trained on USPTO-50k in approximately 5 minutes on a V100.}
\textbf{Metrics.} We measure performance for 
template-relevance prediction using 
template top-$\mathrm{k}$ accuracy, and use reactant top-$\mathrm{k}$ accuracy
for evaluating single-step retrosynthesis 
as described in Section~\ref{sec:background}.

\subsection{Template relevance prediction: USPTO-sm and USPTO-lg\label{sec:template_relevance}}
In this section, we investigate 
template-relevance prediction models in detail.
We first compare the performance of our method
to previously suggested ones. Then we analyze how performance varies
as a function of template frequency.
Next we show that our method can make use of a larger fraction of the training data than a baseline method.
We conclude the section with an ablation study showing
the importance of different modelling parameters.

We analyze the differences
between our \textbf{MHN} method and two related
previously suggested ones:  
Firstly, a fully-connected
network with a softmax output, \textbf{DNN}, in which each output
unit corresponds to a reaction template similar to
\citep{segler2017neuralsymbolic}.
Secondly, \textbf{DNN+pretrain}, which is the same as the above but including a pretraining-step 
on template applicability \citep{fortunato2020data}.
The main differences between these and our MHN method 
are the choice of network architecture (DNN/MHN), 
whether the FPF is used and
whether the pretraining step on applicability prediction is included.
This results in eight possible method 
variants which we include in our studies.
Additionally, we report the results for
a naive baseline \textbf{Pop+FPF}, 
which ranks templates by their popularity 
in the training set and applies the FPF.
For all methods hyperparameters were adjusted on
the validation set as described in 
Appendix~Section~\ref{app:experiments}.

\begin{table}
\caption{Template top-$\mathrm{k}$ accuracy (\%) of different method variants
on USPTO-sm and USPTO-lg. "Model" indicates how the templates were ranked. 
"Filter" specifies if and how templates were excluded from the ranking
via FPF or an applicability check (App).
Pre-train indicates whether a model was pre-trained on the applicability task. Error bars represent confidence intervals 
on binomial proportions.
The grey rows indicate methods specifically proposed here or in prior work.
\label{tab:accuracy}}
\centering
\begin{threeparttable}
\definecolor{light-gray}{gray}{0.95}
\begin{tabular}{llllllllll}
\toprule
\multicolumn{4}{c}{} & \multicolumn{3}{c}{USPTO-sm} & \multicolumn{3}{c}{USPTO-lg} \\
\cmidrule(lr{1em}){5-7}
\cmidrule(lr{1em}){8-10}
                             Ref. & Model & Filter & Pretrain &              Top-1 &             Top-10 &            Top-100 &              Top-1 &             Top-10 &            Top-100 \\
\midrule
\rowcolor{light-gray} \citep{segler2017neuralsymbolic} &   DNN &      - &       no &           $38.1$\tnote{a} &           $64.1$\tnote{a} &           $76.5$\tnote{a} &           $16.0$\tnote{b} &           $35.7$\tnote{b} &           $50.7$\tnote{b} \\
        \rowcolor{light-gray}\citep{fortunato2020data} &   DNN &      - &      yes &           $38.5$\tnote{a} &           $69.1$\tnote{a} &           $85.8$\tnote{a} &           $20.8$\tnote{b} &           $41.7$\tnote{b} &           $54.2$\tnote{b} \\
                                  &   DNN &     FPF &       no &           $39.0$\tnote{a} &           $67.6$\tnote{a} &           $84.6$\tnote{a} &           $17.1$\tnote{b} &           $38.1$\tnote{b} &           $53.6$\tnote{b} \\
                                  &   DNN &     FPF &      yes &           $38.9$\tnote{a} &           $71.2$\tnote{a} &           $90.6$\tnote{a} &  $\mathbf{21.5}$\tnote{b} &           $43.0$\tnote{b} &           $56.0$\tnote{b} \\
                                  &   MHN &      - &       no &           $39.9$\tnote{a} &           $75.7$\tnote{a} &           $91.9$\tnote{a} &           $16.7$\tnote{b} &           $43.6$\tnote{b} &           $71.4$\tnote{b} \\
                                  &   MHN &      - &      yes &  $\mathbf{40.4}$\tnote{a} &           $76.2$\tnote{a} &           $91.8$\tnote{a} &           $16.7$\tnote{b} &           $43.5$\tnote{b} &           $71.4$\tnote{b} \\
\rowcolor{light-gray}(ours)                                &   MHN &     FPF &       no &  $\mathbf{40.5}$\tnote{a} &  $\mathbf{78.7}$\tnote{a} &  $\mathbf{95.9}$\tnote{a} &           $16.9$\tnote{b} &  $\mathbf{44.2}$\tnote{b} &  $\mathbf{72.4}$\tnote{b} \\
                                  &   MHN &     FPF &      yes &  $\mathbf{41.3}$\tnote{a} &  $\mathbf{78.8}$\tnote{a} &  $\mathbf{95.7}$\tnote{a} &           $17.0$\tnote{b} &  $\mathbf{44.1}$\tnote{b} &  $\mathbf{72.3}$\tnote{b} \\
\midrule
                                 &   Pop &      - &       no &           $\enspace 0.0$\tnote{a} &           $\enspace 8.6$\tnote{a} &           $28.9$\tnote{a} &           $\enspace 0.1$\tnote{b} &           $\enspace 0.8$\tnote{b} &           $\enspace 3.5$\tnote{b} \\
                                 &   Pop &     FPF &       no &           $\enspace 1.5$\tnote{a} &           $17.6$\tnote{a} &           $53.1$\tnote{a} &           $\enspace 0.3$\tnote{b} &           $\enspace 1.9$\tnote{b} &           $\enspace 7.5$\tnote{b} \\
                                 &   Pop &    App\tnote{c} &       no &           $\enspace 9.4$\tnote{a} &           $39.6$\tnote{a} &           $80.3$\tnote{a} &           $\enspace 1.1$\tnote{b} &           $\enspace 5.1$\tnote{b} &           $16.5$\tnote{b} \\
                                  
\bottomrule
\end{tabular}

\begin{tablenotes}
\item [a] width of 95\%-confidence interval $<1.3$ \%.  %
\item [b] width of 95\%-confidence interval $<0.4$ \%. %
\item [c] note that the applicability filter violates the modelling constraints given in Section~\ref{sec:background}.
\end{tablenotes}
\end{threeparttable}
\end{table}

\textbf{Comparison to previous methods.}
We compared the predictive performance of
our MHN approach to two closely related 
previously suggested template-relevance prediction 
methods (DNN/DNN+Pretrain) and the popularity baseline (Pop)
on the USPTO-sm/lg data sets described above.
The grey rows in Table~\ref{tab:accuracy} show that for most of the
investigated settings our method exhibits better performance than
these related methods, sometimes by a large margin. Only for top-$1$ accuracy
on USPTO-lg the DNN model with FPF or pre-training outperformed the
MHN model. %

\begin{figure*}
    \centering
    \includegraphics[width=\textwidth]{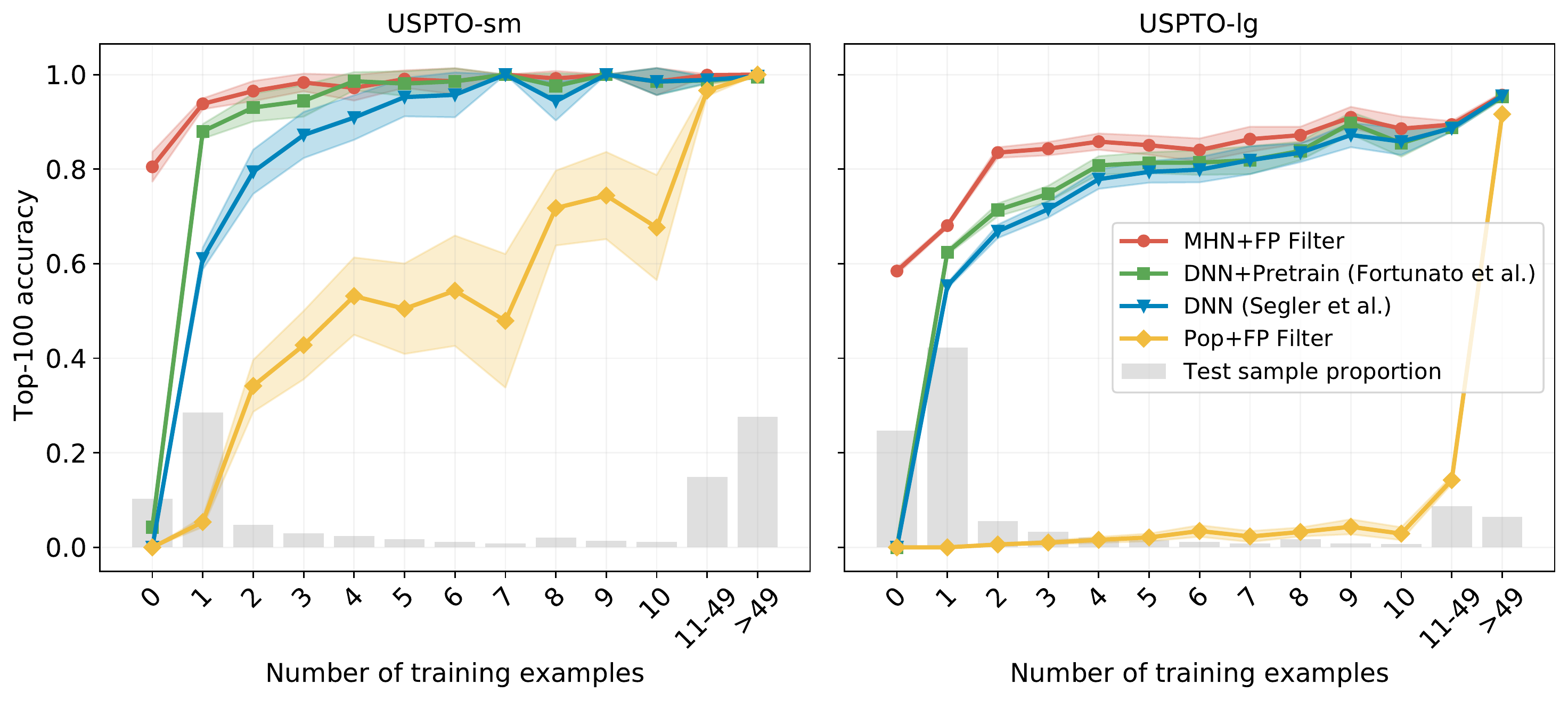}
	\caption{Top-$100$ accuracy for different template popularity on the USPTO-sm/lg datasets. 
	The grey bars represent proportion of samples in the test set.
	Error bars represent 95\%-confidence intervals on binomial proportion.	Our method performs especially well on samples with reaction templates with few training examples.
	\label{fig:nte_reduced}} 
\end{figure*}

\textbf{Rare templates: few- and zero-shot learning.}
Next we investigated how the predictive performance 
for samples varies as a 
function of number of training samples with the same template.
We again compared the methods introduced in the paragraph above.
Figure \ref{fig:nte_reduced} shows the top-$100$ accuracy for different
subsets of the test set, which were grouped by the number of
training samples with the same template.
Especially for samples with rare templates the performance gap
between our method and the compared ones is large.
The DNN models and the popularity baseline perform very poorly for
samples with templates not seen during training, which is expected as they cannot generalize across templates.
The MHN model on the other hand achieves far above random accuracy on
these samples.
The performance on samples with rare templates has a considerable
impact on overall performance due to the large number of templates 
that occur only few times in the
training set. 

\textbf{Learning from rare templates.}
Next we analyze the effect on performance of rare template samples in the training set.
In a classification setting, it is only useful to 
include classes if they are recurring, i.e. represented by more than one sample.
However, in the USPTO-sm/lg datasets many templates occur only once (see Fig.~\ref{fig:template_dist}).
If the templates are modelled as categories as done in the DNN
approach, a large fraction of samples cannot contribute to
performance. 
However, this does not hold for models that can 
generalize across templates, as our MHN model is able to do.
To show the effect of the rare template samples on 
learning, we use the following experiment on USPTO-sm:
We removed all samples with templates that are \emph{exactly once} in the training set and \emph{not} in the test set and retrain the best DNN and MHN
models of the template relevance prediction experiment.
After removal of these samples, the top-$10$ accuracy rose 
from $71.1{\pm}.2$ to $72.3{\pm}.2$
for the DNN+pretrain model and dropped from $78.8{\pm}.4$ to $73.7{\pm}.3$ for the MHN model.
As expected, the performance does not drop for the DNN model, but even improved 
marginally, which we attribute to the model knowing which templates do not occur in the test set.
In contrast, the performance for the MHN model decreased.
This shows that the increased performance of our approach is in 
part caused by the larger fraction of data that can be leveraged for learning.

\textbf{Ablation study.}
\label{sec:ablation}
Next, we investigated the importance of the main 
parameters of the methods reported in this section, 
concretely the choice of network architecture (DNN/MHN),
whether the FPF is used, and
whether pre-training on applicability is performed. 
We evaluated the eight possible parameter
variations on USPTO-sm/lg.
The upper section of Table~\ref{tab:accuracy}
shows the predictive performance of these methods.
We observe that the columns are approximately
sorted from top-to-bottom. In combination with the
organization of the method columns, this implies
that MHN contributes the most to improved
performance, followed by FPF and the use of
pre-training. 
The lower part of Table \ref{tab:accuracy} shows 
the performance of the popularity baseline.
The last row shows that a plain applicability check
is not sufficient for high performance.
We include additional results in
Appendix~Section~\ref{app:experiments}.

\subsection{Single-step retrosynthesis: USPTO-50k\label{sec:retrosynthesis}}
Next we compare our method to 
previously suggested ones in the
single-step retrosynthesis task using the USPTO-50k data set.
We follow the preprocessing procedure of \citep{coley2017computerassisted}
and used rdchiral \citep{coley2019rdchiral} to extract reaction templates.
Following \citep{coley2017computerassisted} 
we shuffled the data and 
assign 80/10/10\% of the samples in each reaction class into 
train/validation/test set respectively.
This is similar to USPTO-sm above but varies in
details discussed in Appendix~Section~\ref{app:experiments}.
We first compare the predictive performance of our
method to previous ones and then investigate its
inference speed.

\begin{table}
    \centering
    \caption{Reactant top-$\mathrm{k}$ accuracy (\%) on USPTO-50k retrosynthesis. Bold values indicate values within 0.1, green 1 and yellow within 3 percentage points to the maximum value. Error bars represent standard deviations across five re-runs. Category ("Cat.") indicates whether a method is template-based (tb) or template-free (tf). 
    Methods in the upper part have been (re-)implemented in this work.
    \label{tab:retroacc}}
    \begin{tabular}{lllllllll}
\toprule
      Abbr. &                              Ref. & Cat. &                                Top-1 &                                Top-3 &                                       Top-5 &                                      Top-10 &                                      Top-20 &                               Top-50 \\
\midrule

  MHNreact &  ours                                 &   tb &                      ${50.5^{\pm.3}}$ &    \cellcolor{jku_green!20}$73.9^{\pm.3}$ &  \cellcolor{jku_green!20}$\mathbf{81.0^{\pm.1}}$ &  \cellcolor{jku_green!20}$\mathbf{87.9^{\pm.2}}$ &  \cellcolor{jku_green!20}$\mathbf{92.0^{\pm.1}}$ &    \cellcolor{jku_green!20}$\mathbf{94.1^{\pm.0}}$ \\
 
  Neuralsym &    \citep{segler2017neuralsymbolic}                               &   tb &                             $45.2^{\pm.2}$ & $67.9^{\pm.5}$ & $75.8^{\pm.2}$ & $83.5^{\pm.2}$ & \cellcolor{jku_yellow!20}$89.1^{\pm.1}$ & \cellcolor{jku_green!20}$93.5^{\pm.1}$
 \\
  Pop &    &   tb &                             $18.4$ & $38.7$ & $48.6$ & $63.0$ & $75.8$ & $89.8$
 \\
\midrule
    Dual-TB &        \citep{sun2020energybased} &   tb &  \cellcolor{jku_green!20}$\mathbf{55.2}$ &  \cellcolor{jku_green!20}$\mathbf{74.6}$ &                  \cellcolor{jku_green!20}$80.5$ &                  \cellcolor{jku_green!20}$86.9$ &                                             &                                      \\
    Dual-TF &        \citep{sun2020energybased} &   tf &          \cellcolor{jku_yellow!20}$53.3$ &                             ${69.7}$ &                                    ${73.0}$ &                                    ${75.0}$ &                                             &                                      \\
     ATx100 &         \citep{sacha2020molecule} &   tf &          \cellcolor{jku_yellow!20}$53.5$ &                                      &         \cellcolor{jku_green!20}$\mathbf{81.0}$ &                 \cellcolor{jku_yellow!20}$85.7$ &                                             &                                      \\
        GLN &    \citep{tetko2020stateoftheart} &   tb &          \cellcolor{jku_yellow!20}$52.5$ &                             ${69.0}$ &                                    ${75.6}$ &                                    ${83.7}$ &                 \cellcolor{jku_yellow!20}$89.0$ &          \cellcolor{jku_yellow!20}$92.4$ \\
 RetroPrime &     \citep{dai2020retrosynthesis} &   tf &                             ${51.4}$ &           \cellcolor{jku_green!20}$74.0$ &                                    ${74.0}$ &                                    ${76.1}$ &                                             &                                      \\
        G2G &        \citep{wang2021retroprime} &   tf &                             ${48.9}$ &                             ${67.6}$ &                                    ${72.5}$ &                                    ${75.5}$ &                                             &                                      \\
      MEGAN &              \citep{shi2020graph} &   tf &                             ${48.6}$ &          \cellcolor{jku_yellow!20}$72.2$ &                  \cellcolor{jku_green!20}$80.3$ &                  \cellcolor{jku_green!20}$87.6$ &                  \cellcolor{jku_green!20}$91.6$ &  \cellcolor{jku_green!20}$\mathbf{94.2}$ \\
    GET-LT1 &          \citep{mao2020molecular} &   tf &                             ${44.9}$ &                             ${58.8}$ &                                    ${62.4}$ &                                    ${65.9}$ &                                             &                                      \\
  Neuralsym &  \citep{segler2017neuralsymbolic,dai2020retrosynthesis} &   tb &                             ${44.4}$ &                             ${65.3}$ &                                    ${72.4}$ &                                    ${78.9}$ &                                    ${82.2}$ &                             ${83.1}$ \\
      GOPRO &    \citep{mann2021retrosynthesis} &   tf &                             ${43.8}$ &                             ${57.2}$ &                                    ${61.4}$ &                                    ${66.6}$ &                                             &                                      \\
      SCROP &       \citep{zheng2020predicting} &   tf &                             ${43.7}$ &                             ${60.0}$ &                                    ${65.2}$ &                                    ${68.7}$ &                                             &                                      \\
   LV-Trans &          \citep{chen2019learning} &   tf &                             ${40.5}$ &                             ${65.1}$ &                                    ${72.8}$ &                                    ${79.4}$ &                                             &                                      \\
      Trans &     \citep{karpov2019transformer} &   tf &                             ${37.9}$ &                             ${57.3}$ &                                    ${62.7}$ &                                             &                                             &                                      \\
   Retrosim &         \citep{coley2017retrosim} &   tb &                             ${37.3}$ &                             ${54.7}$ &                                    ${63.3}$ &                                    ${74.1}$ &                                    ${82.0}$ &                             ${85.3}$ \\
\bottomrule
\end{tabular}

\end{table}

\textbf{Predictive performance.}
Table~\ref{tab:retroacc} shows the reactant top-k
accuracies for different methods.
These methods include, among others, transformer-based \citep{tetko2020stateoftheart,sacha2020molecule},
graph-to-graph \citep{wang2021retroprime} or template-based ones \citep{dai2020retrosynthesis}. 
Some methods \citep{yan2020retroxpert,somnath2020learning,lee2021retcl,guo2020bayesian,liu2017retrosynthetic,ishiguro2020data,hasic2021singlestep, ucak2021substructure} that also report results on USPTO-50k were ommitted here, either due to test set leakage or a different evaluation conditions,
as detailed Appendix~Section~\ref{app:experiments}.
We re-implemented and improved the NeuralSym method as described in Appendix~Section~A3.
and added the popularity baseline described in 
Section~\ref{sec:template_relevance}. 
Hyperparameter selection on the validation set
returned an MHN model with two stacked
Hopfield layers, which we refer to as
\emph{MHNreact} (see Appendix~Section~\ref{app:experiments}).
We ranked reactant sets by the score of the template
used to produce them. If a template execution 
yielded multiple results all were included in the 
prediction in random order. 
Table~\ref{tab:retroacc} shows reactant top-$\mathrm{k}$
accuracy for different methods. 
Our method achieved state-of-the-art performance for $\mathrm{k} \geq 5$ and approaches it for $\mathrm{k}=1,3$. 
Together with Dual-TB \citep{sun2020energybased} this puts 
template-based methods ahead of other approaches at all considered values of k. 

\begin{figure}
    \centering
    \includegraphics[width=\textwidth]{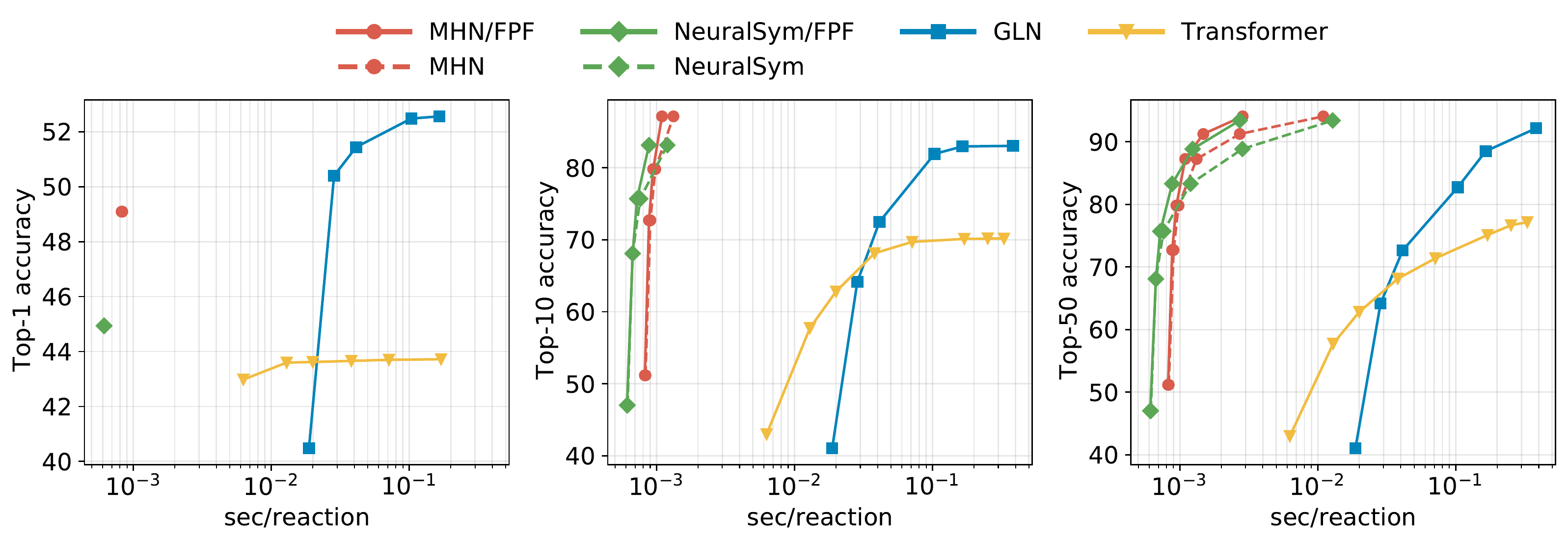}
    \caption{Reactant top-$\mathrm{k}$ accuracy vs. inference speed for different values of $\mathrm{k}$. Upper left is better.
    For Transformer/GLN the points represent different beam sizes. For MHN/NeuralSym the points reflect different numbers of generated reactant sets, namely $\{1,3,5,10,20,50\}$. 
    \label{fig:speed}}
\end{figure}

\textbf{Inference speed.\label{sec:speed}}
Aside from predictive performance, 
inference speed is also vital for retrosynthesis methods.
Therefore, CASP methods are often 
evaluated by their ability to find a route 
in a given time \citep{segler2018planning,chen2020retro,bjerrum2020artificial}.
Template-based methods are sometimes reported to be slow
\citep{shi2020graph,sacha2020molecule}, however
we found that inference speed was not reported in mentioned studies and are seldom reported in general, despite their importance.
Accuracy can be traded for inference speed for many models.
For some, this trade-off is achieved 
by varying beam size 
\citep{dai2020retrosynthesis,tetko2020stateoftheart}. 
In template-based approaches, the number of executed templates can be varied and traded off against speed. 
We compared inference speed of our \textbf{MHN} method with the following baselines. We obtained results for \textbf{GLN} from their paper \citep{shi2020graph}.
We trained a \textbf{Transformer} baseline using the code of \citep{schwaller2019molecular}, as a representative of transformer-based methods \citep{tetko2020stateoftheart,karpov2019transformer,chen2019learning}.
Additionally, we also include the \textbf{NeuralSym}  \citep{segler2017neuralsymbolic} model we implemented in the comparison.
The results are displayed in Fig.~\ref{fig:speed}. 
At comparable or better performance, our method
achieves inference speed of up to two magnitudes
faster compared to Transformer and GLN. 
While NeuralSym is faster than our model 
for some fixed values of accuracy,
MHN yields better maximum accuracy with comparable speed.

\section{Discussion and conclusion} 
\label{sec:discussion}
We have introduced a new Deep Learning architecture
for reaction template prediction based on 
modern Hopfield networks. To this end, 
we have adapted modern Hopfield networks to
associate objects of different modalities and to use a static memory.
This type of multi-modal learning might be relevant in other
areas, in which 
two or more data modalities have to be associated
based on limited available data. 
The architecture comprises a molecule and reaction template 
encoder networks.
The latter enables generalization across templates which enables 
zero-shot learning and improves few-shot learning.
On the single-step retrosynthesis benchmark USPTO-50k, 
our model MHNreact reaches state-of-the-art at
top-$\mathrm{k}$ accuracy for $\mathrm{k}\geq5$.
We found little data that back the common claim of
template-based methods being slow and hope that 
we contributed data that shed a new light on this matter.
We envision that our approach will be used 
to improve CASP systems or synthesis-aware 
generative models \citep{bradshaw2019model, gottipati2020learning, horwood2020molecular, renz2019failure}.\

\emph{Limitations.} 
Our experiments are currently limited by several
factors. 
We did not investigate the 
importance of radius around the reaction center used for
template extraction. We currently do not re-rank
reactants based on a secondary model such as an in-scope
filter \citep{segler2018planning} or dual models
\citep{sun2020energybased}, which could increase
performance. The scalability of our method 
remains to be investigated. 
The results for inference speed depend 
highly on implementation and may
potentially be improved by relatively simple
means. \\
\emph{Potential biases, consequences of failure
and societal impact.} 
Since the inputs to our model 
do not contain data of individuals, there are no direct biases or
unfairness towards a particular gender, age or ethnic groups.
However, there are still biases towards particular chemical
spaces or reactions that have been introduced by humans. 
Our approach can help to 
speed up the discovery of molecules or to find new, and 
more efficient synthesis routes. 
While new molecules
can serve as medicines, 
secure the world’s food supply via agrochemicals, 
or enable a sustainable energy conversion and 
storage to counter or mitigate climate change, 
such new molecules might also be toxins or pollutants, 
which negatively impact humanity. 
As common with methods in machine learning, 
potential danger lies in the possibility 
that users rely too much on 
our new approach and use it without reflecting on the outcomes.
However, the full pipeline in which our method 
would be used includes wet lab tests after its application
to verify and investigate the results, 
gain insights, and possibly derive treatments.
Failures of the proposed method would lead 
to adverse outcomes in wet lab tests.

\section*{Acknowledgements}
The ELLIS Unit Linz, the LIT AI Lab, the 
Institute for Machine Learning,
are supported by
the Federal State Upper Austria.
IARAI is supported by Here Technologies.
We thank the projects
AI-MOTION (LIT-2018-6-YOU-212),
AI-SNN (LIT-2018-6-YOU-214),     
DeepFlood (LIT-2019-8-YOU-213),
Medical Cognitive Computing Center (MC3),
PRIMAL (FFG-873979),
S3AI (FFG-872172),
DL for granular flow (FFG-871302),
ELISE (H2020-ICT-2019-3 ID: 951847),
AIDD (MSCA-ITN-2020 ID: 956832).
We thank
Janssen Pharmaceutica (MaDeSMart, HBC.2018.2287),
Audi.JKU Deep Learning Center,
TGW LOGISTICS GROUP GMBH,
Silicon Austria Labs (SAL),
FILL Gesellschaft mbH,
Anyline GmbH,
Google,
ZF Friedrichshafen AG,
Robert Bosch GmbH,
UCB Biopharma SRL,
Merck Healthcare KGaA,
Software Competence Center Hagenberg GmbH,
T\"{U}V Austria,
and the NVIDIA Corporation.

\printbibliography

\appendix

\newpage
\section*{APPENDIX}
\tableofcontents

\setcounter{figure}{0}    
\setcounter{table}{0}    

\renewcommand{\thesection}{A\arabic{section}}
\renewcommand{\thefigure}{A\arabic{figure}}
\renewcommand{\thetable}{A\arabic{table}}

\newpage
\section{Notation}

\begin{table*}[htp]
    \centering
    \begin{tabular}{lcl} 
        \toprule
        Definition &  Symbol/Notation & Dimension \\ 
        \midrule
         set of reaction templates       & $\BT$  & set of size $K$ \\
         encoded set of reaction template& $\BT_h$  & $d_t \times K$\\
         reactant molecules              & $\Br$ & \\
         product molecule                & $\Bm$ & \\
         encoded molecule                & $\Bm_h$ & $d_m$ \\
         reaction template               & $\Bt$ or $\Bt_k$ & \\
         training set pair               & $(\Bm,\Bt)$ & \\
         state pattern                   & $\Bxi$ & $d$ \\
         stored pattern                  & $\Bx_k$        & $d$   \\
         stored pattern matrix           & $\BX$          & $d \times K$ \\  associations                    & $\Bp$     &  $K$  \\
         update function of MHN          & $f$            &  \\
         molecule encoder                & $\Bh^m$            &  \\
         reaction template encoder       & $\Bh^t$            &  \\
         model function                  & $g$            &  \\
         network parameters of $\Bh^m$     & $\Bw$      &  \\
         network parameters of $\Bh^t$     & $\Bv$      &  \\
         parameters of Hopfield layer $\Bh^m$     & $\BW_{m}, \BW_{t}$      &  $d \times \text{undef}$\\
         association activation function & $\phi$            &  \\
         number of templates             & $K$       & \\
         dimension of association space  & $d$       & \\
         \bottomrule
    \end{tabular}
    \caption{Symbols and notations used in this paper.}
    \label{tab:notation}
\end{table*}

\clearpage

In this appendix, we first provide more
related work (see Section~\ref{app:related_work}), 
then details on the 
experiments (see Section~\ref{app:experiments}),
additional results (see Section~\ref{app:results}),
visualization (see Section~\ref{app:hopfield_space}),
alternative view on the loss and an 
extended formulation of the algorithm
as pseudo-code (see Section~\ref{app:loss}).

\section{Further related work}
\label{app:related_work}
Here we provide a broader view on works that 
have addressed common issues with template 
relevance prediction.
In prior work \citep{baylon2019enhancing, ishida2019prediction, fortunato2020data, segler2017neuralsymbolic, segler2018planning, bjerrum2020artificial}, template relevance prediction is often viewed as a multi-class classification task, where, given a product, an ML model is trained to predict which of the templates extracted from a reaction database are most relevant. 
Automatic extraction of templates leads to many rare templates, which poses a problem in the classification task as it leads to many classes with few training samples \citep{fortunato2020data, segler2018planning}.
In earlier work \citep{segler2018planning}, rare templates were excluded from training.
\citet{baylon2019enhancing} proposed a hierarchical grouping of reaction templates and trained a separate NN for each group of templates.
\citet{fortunato2020data} pretrained their template scoring model to predict which templates are
applicable and observed that it improves template-relevance
predictions, especially for rare templates.
\citep{ishida2019prediction} used a graph-NN to predict the relevant templates.
\citet{bjerrum2020artificial} trained two separate NNs. The first NN is trained on
the applicability matrix and serves for pre-filtering reaction
templates. The second is trained on the reaction dataset, and
ranks the reaction templates according to their relevance.
\citet{dai2020retrosynthesis} make use of the template structures. However, they factorize the predicted probabilities into multiple functions, which might not be suited to the problem.
The in-scope filter of the CASP system by
\citet{segler2018planning} also make use of 
template structure.
\citet{sun2020energybased} apply all templates and uses a model to rank the resulting reactants, which achieves the best top-1 accuracy but is computationally costly (Dual-TB in Table~\ref{tab:retroacc}).

\section{Details on Experiments}
\label{app:experiments}
\subsection{Template relevance prediction}

\subsubsection{Datasets and preprocessing}
\label{sec:datasets}

For preprocessing USPTO-sm and USPTO-lg, 
we followed the implementation of \citet{fortunato2020data}.
The templates were extracted from the 
mapped reactions using RDChiral \citep{coley2019rdchiral} and 
subsequently filtered according to symmetry, validity, and by 
checking if the application of the template yielded the 
result as in the reaction the template originated from.
Despite adhering to the original implementation by the authors,
our preprocessing resulted in different dataset sizes.
The roughly 2 million starting reactions decreased to 443,763 samples and 236,053 reaction templates
for USPTO-lg (compared to 669,683 samples and 186,822 reaction templates in \citet{fortunato2020data}), 
and to 40,257 samples and 9,162 reaction templates for USPTO-sm 
(compared to 32,099 samples and 7,765 reaction templates in \citet{fortunato2020data}).
It can be seen that 17\% of samples occur in a class that has only one sample in USPTO-sm and 43\% in USPTO-lg. 66\% of reaction templates in USPTO-sm and 80\% in USPTO-lg occur only in a single reaction.

To allow for pretraining of the methods, we calculated the 
applicability matrix, i.e., which templates in a dataset are applicable to which molecules.
\citet{fortunato2020data} reported that this would take  ${\sim}36$ CPU-hours for USPTO-sm and ${\sim}$330 CPU-days for USPTO-lg on a single core of a Xeon(R) Gold 6154 CPU@3 GHz.
We found that using a substructure screen could speed up this
procedure. The following values were obtained using an AMD EPYC 7542. For USPTO-sm it only takes us 3.3 CPU-minutes to achieve
the same result. Using 16 CPU-cores this reduces to ${\sim}14s$.
Using 32 CPU-cores applicability calculation time for USPTO-lg
reduces to ${\sim}50m$ which corresponds to 27 CPU-hours compared to ${\sim}$8000 CPU-hours for the original implementation.
These comparisons should be taken with a grain of salt because of the slightly different dataset sizes and hardware used.
For USPTO-lg $443 \ 763 \cdot 236 \ 053 \sim 10^{11}$ pairs have to be checked, and our code relies on a python loop.
Using a compiled language could probably further speed up the procedure.

\subsubsection{Data splits.}
We split the data into a training, validation, and test set 
following \citet{fortunato2020data}. Here, a stratified split was
used to ensure that templates are more equally represented 
across the splits.
Concretely, in \citet{fortunato2020data},
the split proportions were 80/10/10\% except for templates
with  fewer than 10 samples, where one random sample was 
put into the test set, one into the validation set, and the rest
into the training set. If only two samples were present,
one was put into the test and one into the training set.
Finally, if only a single sample was available for a template, it was
randomly placed into the train/validation/test set with an
80/10/10\% chance.

\subsubsection{Feature extraction.\label{sec:featurization}}
\textbf{Molecule fingerprints.} The source molecules are represented as SMILES. We extract a fingerprint 
representation of the molecules. 
We tried out different fingerprint types, 
e.g. folded Morgan fingerprints with chirality \citep{morgan1965generation}
and the hyperparameter selection procedure (see Table~\ref{tab:hparams_tr}) 
selected Morgan fingerprint with radius of 2 folded to 4096 features. 

\textbf{Template fingerprints.} For the template representation, a similar procedure has been applied. A template consists of multiple enumerated SMARTS-strings. 
The fingerprint type for the templates was set to 'rdk'-fingerprint or 'pattern'-fingerprint for template relevance prediction and 
calculated for each molecule that the pattern represents.
We experimented with multiple ways of combining not just the product side, but 
also the reactant side to this representation.
We found the following to perform best among the considered variants.
The fingerprints were calculated for each molecular pattern, and a disjunction 
over reactants as well as products was calculated. The product minus half of 
the reactant side results in the template fingerprint as input for the 
template encoder. We also tried the 
'structuralFingerprintForReaction' function provided by RDKit, which concatenates the disjunction of each side of the reaction, but found the weighted combination to perform better.
The resulting representation was similar and for some equal, and therefore, we added an additional random template embedding.
The random noise was added to the representation of frequent templates in order to help to discriminate frequent templates with a high fingerprint similarity.
Templates are classified as "frequent" if they appear at least a certain
number of times in the training-set, which is determined by the hyperparameter "random template threshold" (see Table~\ref{tab:hparams_tr}).

\subsubsection{Training}
All models were trained for a maximum of 100 epochs on a Titan V with 12 GB RAM or a P40 with 24 GB RAM using PyTorch 1.6.0 \citep{paszke2019pytorch}.
In the case of DNN, only the molecule encoder was trained, and a linear layer, 
projecting from the last hidden layer to the number of templates was added. For pretraining on the applicability task we changed the loss function 
to the mean of binary cross-entropy for each output (template).  
We also experimented with InfoNCE-loss \citep{oord2018representation} on
representations in Hopfield space (see Appendix~\ref{app:loss}).
Because of fast convergence and slightly better performance, and because 
for the USPTO data sets only a single template is correct for 
each molecule, we use our proposed loss, which in this case 
is equivalent to CE-loss. 

\subsubsection{Hyperparameter selection and model architecture} 
Hyperparameters were explored via automatic Bayesian optimization for USPTO-sm, as well as manual hyperparameter-tuning. In the former, early stopping was employed. The range of values was selected based on prior knowledge. Additional manual hyperparameter-tuning 
resulted in better predictive performance on the validation set.
Some of the important hyperparameters are the beta scaling factor of the Hopfield layer $\beta$, the dimension of the association space $d$, as well as the association-activation function, or if the association space should be normalized via layer-norm \citep{ba2016layer}.
An overview of considered and selected hyperparameters is given in Tab.~\ref{tab:hparams_tr}. All models were trained if applicable for a maximum of $100$ epochs using AdamW\citep{loshchilov2018fixing} (betas=(0.9, 0.999), eps=$ 1\mathrm{e}{-8}$, weight\_decay=$1\mathrm{e}{-2}$, amsgrad=False).
Hyperparameters were selected based on the minimal CE-loss on the validation set.

\begin{table*}
\centering
\begin{tabular}{lccc}
\hline 
~ & ~ & \textbf{MHN selected} & \textbf{DNN selected}\\
\textbf{Hyperparam} & \textbf{Values}  & \textbf{(Sm/Lg)} & \textbf{(Sm/Lg)}\\
\hline 
learning rates & $\{ 1\text{e-}4, 2\text{e-}4, 5\text{e-}4 , 1\text{e-}3 \}$ & $5\text{e-}4$ / $1\text{e-}4$ & $5\text{e-}4$ / $2\text{e-}4$ \\
batch-size & $\{32, 128, 256, 1024\}$ & $1024$ & $256$ / $1024$\\ 
dropout & $[0.0, 0.6]$ & $0.2$ & $0.15$ \\
\midrule
\emph{molecule encoder} & & \\
fingerprint type & \{morgan, rdk\} & morgan & morgan \\
fingerprint size & $\{1024, 2048, 4096\}$ & $4096$ & $4096$\\
number of layers & \{0, 1, 2\}& 0 & 1 \\
layer-dimension & \{1024, 2048, 4096\}& - & 2048 \\
activation-function & \{None, SELU, ReLU\} & None & ReLU \\
\midrule
\emph{template encoder} & & \\
number of layers & \{0, 1, 2\}& 0 \\
template fingerprint type & \{pattern, rdk\} & rdk \\
random template thresh. & {-1, 2, 5, 10, 50} & 2 \\
\midrule
\emph{Hopfield layer} & & \\
beta & $[0.01, 0.3]$ & $0.03$\\
association af & \{None, SELU, GeLU, Tanh\} & None / Tanh\\
normalize pattern & \{False, True\} & False \\
normalize projection & \{False, True\}& True \\
learnable stored-pattern & \{False, True\} & False \\
hopf-num-layers & \{1, 2, 3\}& 1\\
hopf-num-Wm & \{1, 2, 3\} & 1\\
hopf-num-Wt & \{1, 2, 3\} & 1\\
hopf-FF-activation & \{None, SELU, ReLU\} & None  & \\
association-dimension  $d$ & \{32, 64, 512, 1024\} & 1024\\
hopf-num-heads & $\{1, 6, 12\}$ & 1\\
\midrule
\emph{Setting-specific-hps} & & \\
pretraining epochs & \{0,5,10,15,20,25\} & 10 & 25 / 5 \\

\hline
\end{tabular}
\caption[Hyperparameters for template relevance prediction]{Hyperparameter search-space for template relevance prediction. 
The rows are subdivided into five modules: overall parameters, the molecule encoder, the template encoder, the Hopfield layer, and setting specific hyperparameters that were only used if explicitly stated. 
The column values show the range of the explored parameters.
If multiple Hopfield layers were used, the same hyperparameters were used for all layers.
A "random template threshold" of -1 corresponds to not adding noise. The fingerprint size for the molecule encoder was always the same as the template encoder. The pretraining learning rate was also defined by the learning rate, the optimizer remained the same, but the loss-function changed to binary-cross-entropy loss. \label{tab:hparams_tr}}
\end{table*}

\subsection{Single-step retrosynthesis}
\subsubsection{Datasets and preprocessing}
For single-step retrosynthesis, we used the preprocessed version and splitting-procedure from \citet{coley2017retrosim}. The dataset originated from USPTO-50k by \citet{schneider2015development}. It is different in details from USPTO-sm and does not contain a filtering step, whereby samples are excluded if extracted and applied templates don't yield the reactants. A further difference is the split.
For USPTO-50k, we shuffle the samples and further split it according to the procedure by \citet{coley2017retrosim}, randomly splitting within the reaction types, to obtain 40008 train- 5001 validation- and 5007 test-samples (80/10/10). We computed the reaction templates only from the train- and validation-set.

\subsubsection{Feature extraction}
\label{sec:lgamma}
For this experiment, we additionally explored Mixed-Fingerprint (MxFP), which is a mixture of multiple unfolded, counted (where applicable) RDKit fingerprints: MACCS, Morgan, ErG, AtomPair, TopolocialTorsion, and RDK. For each fingerprint type, we sort the features by their variance of binary fingerprints in the train-set and discard low-variance features up to a certain length. We additionally scale the counts by $\log(1+x)$ \citep{segler2018planning}.

\emph{Template-representation.}
We compute fingerprints for each subgraph-pattern in the reaction template. Again we use a mixture of multiple unfolded RDKit fingerprints.
For pooling the reactant fingerprints, we additionally experimented with different pooling operations.
The main idea is to avoid that different sets are identical 
after pooling and thus to increase the expressivity 
of the pooling operation.
Lgamma pooling is a novel pooling operation that 
uses the log of the gamma-function. 
\begin{align}
    \text{lgp}(\Bx) =  \log \left( \Gamma \left( \sum_{i=0}^{n}{ x_i +2} \right) \right) - \sum_{i=0}^{n} \log ( \Gamma ( x_i+1)),
\end{align}
where the $\Bx$ contains a single feature
of the elements of the set that is pooled. 
The use of this pooling function provided 
a small performance increase over max-pooling.

\subsubsection{Hyperparameters and model architecture}
Hyperparameters were tuned manually and selected based on top-1 accuracy on the validation set.  The explored parameters, as well as the selected hyperparameters, can be found in Table \ref{tab:hparams_ssr}. Models were also trained if applicable for a maximum of 100 epochs using AdamW \citep{loshchilov2018fixing} (betas=(0.9,0.999), eps=1e-8, weight\_decay=1e-2, amsgrad=False). 
As input, MxFP was selected with a fingerprint size of 30k.
It consists of two layers, where the input for the second layer is $\Bxi^{\mathrm{new}}+\Bxi$, a skip connection from the first layers input plus the output of the first layer. The first layers' memory is comprised of MxFP-template fingerprints with lgamma-pooled reactants (see Section ~\ref{sec:lgamma}). 
The second layer uses a different template representation: RDK-template fingerprint with additional random noise for all templates which appear more than once in the training set.
The final prediction is computed by a weighted average of the individual layers' $\Bp$.

The NeuralSym baseline was trained as follows:
As a model architecture, we used a feed-forward neural network with a single hidden layer of size 4096 and SELU activation function \cite{klambauer2017selfnormalizing}.
The inputs to this network were ECFP-fingerprints \citep{rogers2010extendedconnectivity} with radius 2 and size 4096. The model was trained using AdamW \citep{loshchilov2018fixing} with learning rate 1e-3
and weight-decay of 1e-3. The model was trained for 7 epochs with a batch size of 512.

\begin{table*}
\centering
\begin{tabular}{lccc}
\hline 
\textbf{Hyperparam} & \textbf{Values}  & \textbf{MHN (50k)} & \textbf{DNN (50k)}\\
\hline 
learning rates & $\{ 1\text{e-}4, 2\text{e-}4, 5\text{e-}4 , 1\text{e-}3 \}$ & $5\text{e-}4$ / $1\text{e-}4$ & $1\text{e-}4$ \\
batch-size & $\{32, 128, 256, 1024\}$ & $1024$ & $256$ \\ 
dropout & $[0.0, 0.6]$ & $0.2$ & $0.15$ \\
\midrule
\emph{molecule encoder} & & \\
fingerprint type & \{morgan, rdk, MxFP\} & MxFP & morgan \\
fingerprint size & $\{4096, ..., 40\text{e}3\}$ & $3\text{e}4$ & $4096$\\
fingerprint radius & $\{2, ..., 6\}$ & - & 2 \\
number of layers & \{0, 1, 2\}& 0 & 1 \\
layer-dimension & \{1024, 2048, 4096\}& - & 4096 \\
activation-function (af) & \{None, SELU, ReLU\} & None & SELU \\
\midrule
\emph{template encoder 1} & & \\
number of layers & \{0, 1, 2\} & 0 \\
template fingerprint type & \{rdk, rdkc, MxFP\} & MxFP \\
random template threshold & {-1, 2, 5, 10, 50} & -1 \\
reactant pooling & \{max, sum, mean, lgamma\} & lgamma \\
\midrule 
\emph{template encoder 2} & & \\
number of layers & \{0, 1, 2\} & 0 \\
template fingerprint type & \{rdk, MxFP\} & rdk \\
random template threshold & {-1, 2, 5, 10, 50} & 2 \\
\midrule
\emph{Hopfield layer 1 and 2} & & \\
beta & $[0.01, 0.3]$ & $0.03$\\
association af & \{None, Tanh\} & None \\
normalize input pattern & \{False, True\} & True \\
normalize association proj.& \{False, True\}& True \\
learnable stored-pattern & \{False\} & False \\
hopf-num-layers & \{1, 2\}& 2\\
hopf-num-Wm & \{1, 2\} & 1\\
hopf-num-Wt & \{1, 2\} & 1\\
hopf-FF-af & \{None, SELU, ReLU\} & None  & \\
association-dimension  $d$ & \{32, 64, 512, 1024\} & 1024\\
hopf-num-heads & $\{1, 6, 12\}$ & 1\\

\hline
\end{tabular}%
\caption[Hyperparameters for single-step retrosynthesis]{Hyperparameter search-space for single-step retrosynthesis.
The layout and specifics are equivalent to Table~\ref{tab:hparams_tr} but differ in the explored values and architectural choices. The hyperparameters for the Hopfield layer remain the same among layers, with individually initialized weight parameters. The input for layer 1 is given by "template encoder 1" and vice versa for layer 2. The column MHN (50k) corresponds to the results of MHNreact and DNN (50k) to the first mention of Neuralsym in Table~\ref{tab:retroacc}.
\label{tab:hparams_ssr}}
\end{table*}

\subsubsection{Methods omitted from comparison}
We omitted some studies
from the comparison in Table~2, 
despite them reporting performance on USPTO-50k.
We found that the experimental settings or reported metrics
in these studies differed 
from ours.
While these reported values are not per se flawed, we think
that inclusion in the comparison may be misleading.
We list specifics below:
\begin{itemize}
\item \citet{yan2020retroxpert} reported (\url{https://github.com/uta-smile/RetroXpert}) that their model used information in the atom-mappings about where the reaction center is. This information relies on the knowledge of the reactants. As the reactants are to be predicted in this task this is considered test set leakage.

\item The reported values in \citep{somnath2020learning}
are also based on unintentional use of information about the reaction center, similar to above\footnote{Personal communication with the authors}.

\item The method proposed in \citep{lee2021retcl} selects reactants from a candidate set. Since this candidate set is a superset of the reactants in the USPTO-50k, it might contain information about the test data.
Indeed we found that we could augment the performance of our method by a process of elimination, i.e., discarding reactant sets from the predictions if they are not in the candidate set. 

\item The method proposed in \citep{hasic2021singlestep} also relies on a candidate set that we suspect to contain information about the test set.
However, the description of the method is not very detailed.

\item \citet{guo2020bayesian} use reactants from USPTO-stereo as described in \citep{schwaller2019molecular} as a  candidate set. We found that, given this set, we could 
augment the performance of our method by removing reactant sets not in this set from our predictions.

\item \citet{ishiguro2020data} propose a pretraining step on a larger data set which does not conform to the setting in most prior work and is therefor excluded. 

\item \citet{ishida2019prediction} use a different subset of USPTO-50k to train their model and report different metrics.

\item \citet{ucak2021substructure} also make use of a different subset and also do not report reactant top-k accuracy.

\item \citet{liu2017retrosynthetic} only provide results for the special case where the type of reaction is provided to the model.    
\end{itemize}

\subsubsection{Inference speed}
We investigated the speed/performance trade-off for multiple methods. 
Firstly, we trained a Transformer baseline using the code and settings
provided by \citep{schwaller2019molecular}, except for setting the batch size to 8192,
warmup steps to 6k, and train steps to 50k.
We evaluated the predictions of this model when run with beam sizes $\{1,3,5,10,20,50,75,100\}$. While performance increases with larger beam size, the model also gets slower.
This model outperforms the model suggested in \citep{karpov2019transformer}, but could not
reach the performance of \citep{tetko2020stateoftheart}.
Model training took about six hours on an Nvidia V100.

For MHN and NeuralSym, the inference procedure contains the
following steps. First fingerprints for the given products
have are generated. Then the model is used to predict
template relevance. 
For each product templates are executed in the order
of their score until a fixed number of reactant sets are
obtained. 
To optimize top-k accuracy, it does not help to generate more than k reactant sets. 
Therefore we set the number of reactant sets to generate to
 $\{1,3,5,10,20,50\}$ optimize speed without loss of top-k
 accuracy for the respective $k$ and measured inference time.
Speeds for the Transformer, NeuralSym, and MHN models
have been measured using an Nvidia Tesla T4 and 16 cores of an AMD EPYC 7542.
We also tested stopping template
execution based on the cumulative probability of already
executed ones as done in \citep{segler2018planning},
however found that it did not improve upon stopping
after a certain number of reactants have been retrieved.

\section{Additional Results}
\label{app:results}

\begin{figure*}[htp]
    \centering
    \includegraphics[width=\textwidth]{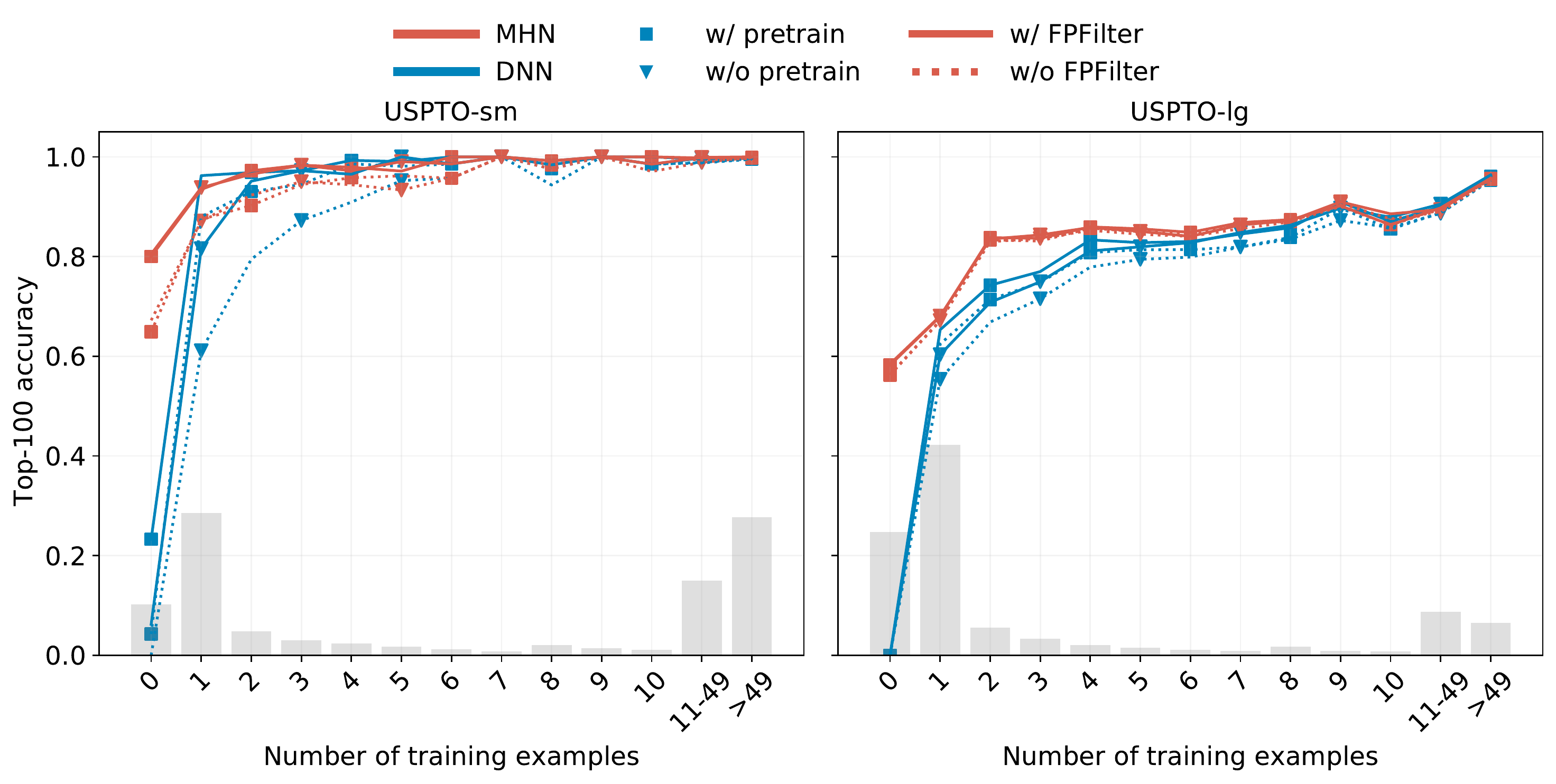}
	\caption[Ablation study and few-shot learning]{Results of methods with different
	design elements	on the USPTO-sm and USPTO-lg datasets.
	Each method consists of a combination of 
	the following elements: 
	a) a network MHN or DNN (blue or red line), 
	b) whether pretraining is applied (squares or triangles), and 
	c) whether fingerprint filter is applied for 
	postprocessing (solid or dashed line). 
	These eight possible combinations
	are displayed as lines with their top-100 accuracy on 
	the y-axis and the different template frequency
	categories on the x-axis.
	\label{fig:nte_all_variations}} 
\end{figure*}
\begin{figure*}[htp]
    \centering
    \includegraphics[width=\textwidth]{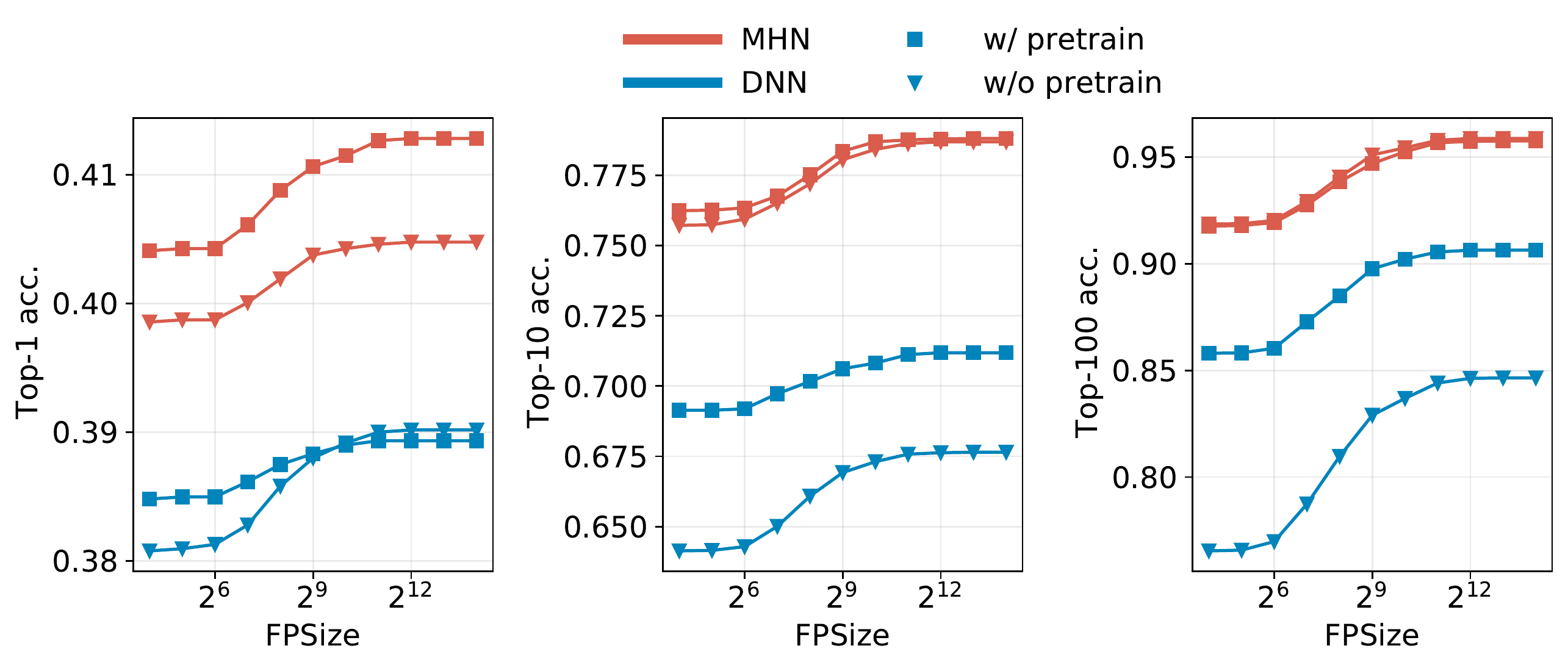}
	\caption[Predictive performance and fingerprint size]{Predictive performance of different 
	methods in dependency of the fingerprint size.
	Each method consists of a combination of 
	the following elements: 
	a) a network MHN or DNN (blue or red line), and
	b) whether pretraining is applied (squares or triangles).
	These four possible combinations
	are displayed as lines with their top-1, top-10 and 
	top-100 accuracy.
	Performance saturates at a fingerprint size of about $2^{12}{=}4096$, and we therefore choose this value for the other experiments.
	\label{fig:result_fp_size}} 
\end{figure*}
\begin{figure*}[htp]
    \centering
    \includegraphics[width=\textwidth]{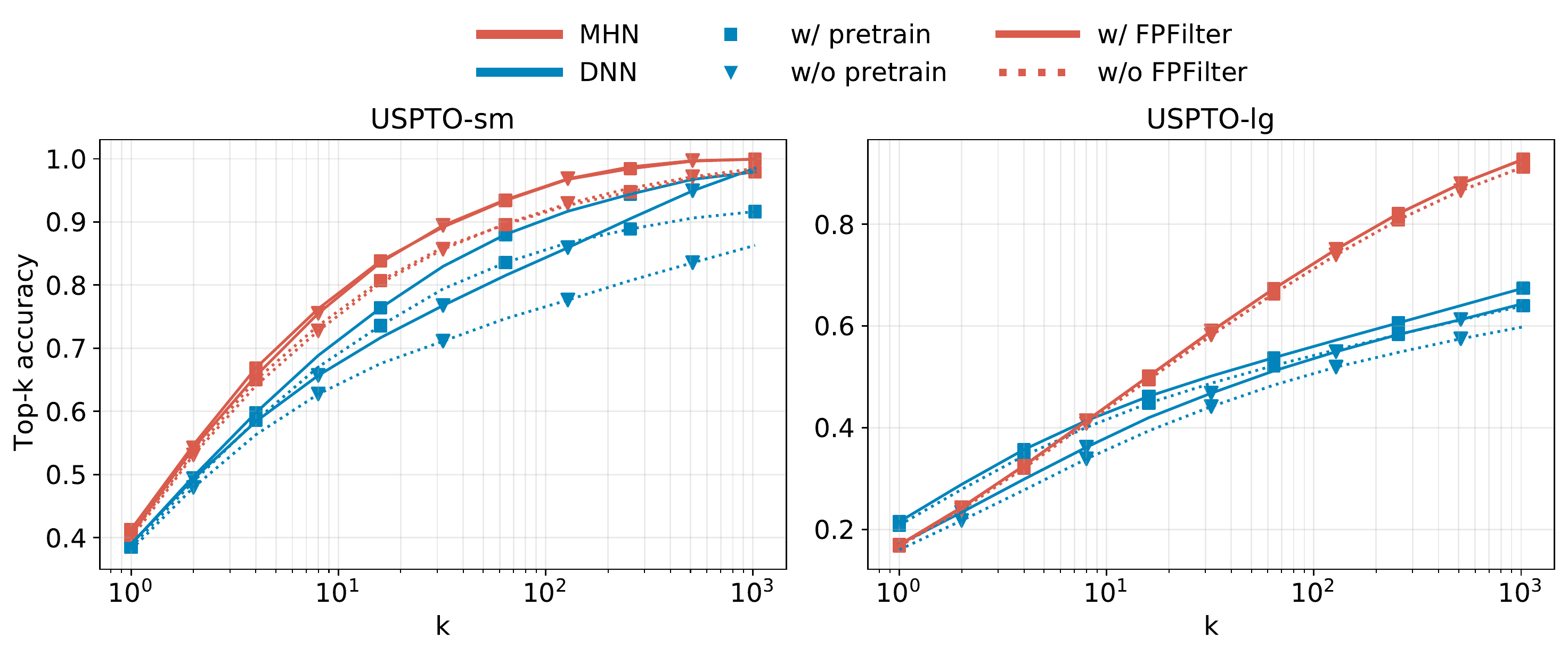}
	\caption[Ablation study and top-k accuracy]{Comparison of methods with respect to
	their top-k accuracies.
	Each method consists of a combination of 
	the following elements: 
	a) a network MHN or DNN (blue or red line), 
	b) whether pretraining is applied (squares or triangles), and 
	c) whether fingerprint filter is applied for 
	postprocessing (solid or dashed line). 
	These eight possible combinations
	are displayed as lines with their k parameter on 
	the x-axis and their top-k accuracy on the y-axis.
	MHNs provide the best top-k accuracy with k larger 
	or equal 10. 
	\label{fig:result_k_topk}} 
\end{figure*}

\clearpage
\section{Hopfield Association Space}
\label{app:hopfield_space}
\begin{figure}[htp]

\centering
\includegraphics[width=1\textwidth]{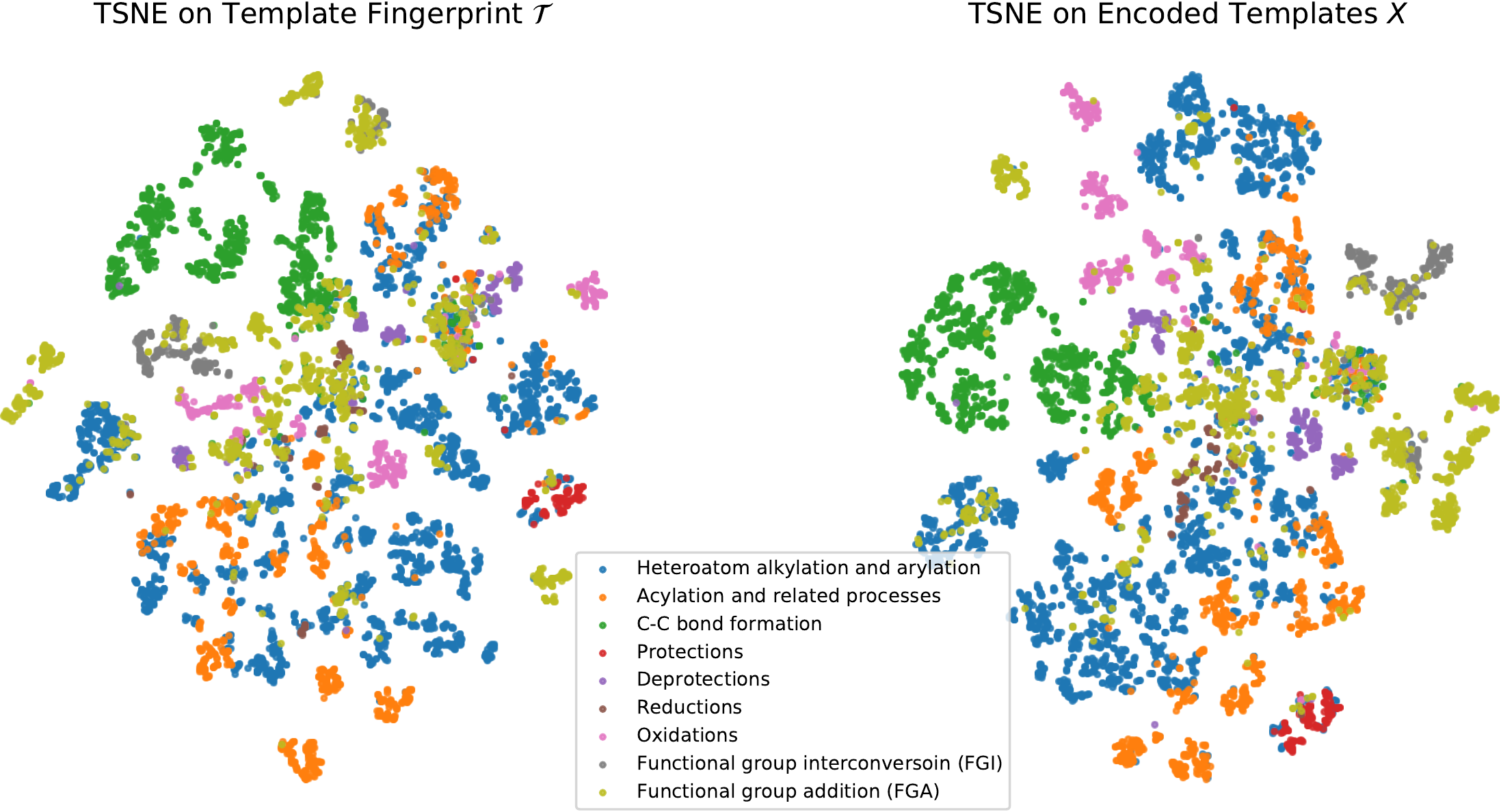}
\caption[Visualization of the learned template representations]{t-SNE downprojection of the reaction template fingerprints (left) and learned representations
of reaction templates $\boldsymbol{X}$ (right). The 
colors represent reaction types of 
substructure-based expert systems as categorized by \citep{schneider2015development}. 
}
\label{fig:TSNE}
\end{figure}

Figure \ref{fig:TSNE} shows a t-SNE %
embedding of both the reaction fingerprints and the learned embeddings.
Each point is colored according to its class as defined in \citep{schneider2015development}.

For example, reactions belonging to the type "oxidations" can be distant in the fingerprint space (pink points in the
left figure), while in their learned representations
are closer (pink points in right figure).
Note that our model did not have access to these reaction types.
It can be seen that the chosen representation for reaction templates already captures information about the relationship, and the same reaction types are represented closer after embedding it using t-SNE.

\clearpage

\section{Objective and loss functions}
\label{app:loss}
\paragraph{Loss on retrieved patterns.}
We provide a more general view on the objective and the loss function 
from the perspective of Hopfield networks and retrieving patterns. 
The main idea is to retrieve patterns from label space, rather than from 
Hopfield space, because the loss functions operate in the label space.  
The last Hopfield layer of our architecture supplies both $\Bp$, the softmax
vector of probabilities of drawing reaction templates, and $\Bxi^{\mathrm{new}}$,
an average of reaction template representations $\Bx$. 
However, averages of reaction templates are no longer reaction templates, 
but we are interested in the {\em probability} $\ell_{\xi}$ 
of drawing a $\Bx_k$ that fits to $\Bxi$.
The probability $\ell_{\xi}$ can still be 
computed via a slightly modified Hopfield network update,
where instead of retrieving from a memory $\BX$ of template representations
in Hopfield space, we retrieve from the space of labels or scores.
Such an update has been introduced previously and uses 
stored patterns that are augmented by labels \citep[p.83ff]{ramsauer2021hopfield}.

The probability $\ell_{\xi}$ of drawing a $\Bx_k$ that 
fits to $\Bxi$ can be computed
by a modified Hopfield network update:
\begin{align}
\label{eq:updateLabel}
  \Bq_{\xi} \ &= \ \BL \ \Bp \ = \   
\BL \ \soft ( \beta \BX^T \Bxi) \ , \ \ 
\ell_{\xi} \ = \ \BOn^T \ \Bq_{\xi} \ ,
\end{align}
where $\BL \in \dR^{K \times K}$  is the diagonal 
matrix of labels that are zero or one and
$\BOn \in \dR^K$ is the vector of ones. 
In general, $\BL$ can be used to include unlabeled 
data points as stored patterns, 
where $\BL$ is a Gram matrix times a diagonal label matrix 
to transfer label information from labeled data points to unlabeled data points.
$L_{kk}=1$ means that $\Bx_k$ fits to $\Bxi$ and 
$L_{kk}=0$ means that $\Bx_k$ does not fit to $\Bxi$.
In the context of the Hopfield networks, if more than one $\Bx_k$ fits to $\Bxi$
then all $\Bx_k$ that fit to $\Bxi$ constitute a metastable state.
Instead of $L_{kk}$ being equal to zero or one, $L_{kk}$ can give a non-negative score
for how well $\Bx_k$ does fit to $\Bxi$.
In this case $\ell_{\xi}$ is the {\em expected score}.

In this general view, our objective is to minimize $-\log(\ell_{\xi})$ .
If $L_{kk}$ is equal to zero or one,
$\log(\ell_{\xi})$ is the log-likelihood of 
drawing a fitting $\Bx_k$.
If only one $\Bx_k$ fits (exactly one $L_{kk}$ is one), 
then our objective is equivalent to 
the cross-entropy (CE) loss for 
multi-class classification. 
However, if more $\Bx_l$ are labeled as fitting, 
then our objective is different from CE, which might not be appropriate.
If $L_{kk}$ is a non-negative score for the molecule-template
pair, our approach will maximize the expected score.

\paragraph{Cross-entropy loss.} 
In a simple setting, in which each molecule only has 
a single correct reaction template in $\BT$, 
a categorical cross-entropy loss is equivalent to
the suggested loss. 
We encode the correct template by a one-hot vector 
$\By=(0,\ldots,0,1,0,\ldots,0)$, 
where $1$ indicates the position of the correct
template in the template set $\Bt=\Bt^k$.
We then minimize the cross-entropy loss function 
$\ell_{\mathrm{CE}}( \By, \Bp)=\mathrm{crossentropy}(\By, \Bp)$
between ground truth $\By$ and the model's 
predictions $\Bp$ for a single pair of the training set and
the overall loss is an average over all such pairs.
The corresponding algorithm is given in Alg.~\ref{alg:mhn_general}.

\paragraph{InfoNCE in Hopfield space.} 
An almost equivalent loss function would 
be to use a contrastive loss on the retrieved pattern $\Bxi^{\mathrm{new}}$.
This contrastive loss measures the cosine similarity
of the retrieved pattern with 
the correct stored patterns with 
the InfoNCE function \citep{chen2020simple,oord2018representation}: 
\begin{align}
    \label{eq:infonce}
    \ell_{\mathrm{p}}(\Bxi^{\mathrm{new}},\Bx^+,\BX^-)=
    \mathrm{InfoNCE} (\Bxi^{\mathrm{new}},\Bx^+,\BX^-)=
    -\log \frac{ \exp(\mathrm{sim}(\Bxi^{\mathrm{new}}, \Bx^+)/\tau)}{\sum_{k} \exp(\mathrm{sim}(\Bxi^{\mathrm{new}}, \Bx^-_k)/\tau)},
\end{align}
where $\mathrm{sim}(.,.)$ is a
pairwise similarity function,
$\Bx^+$ is the
representation of the
correct reaction templates,
and $\BX^-$ is the set of 
representations 
of the incorrect reaction templates, 
that are contrasted against each other.
This loss is equivalent to cross-entropy loss if 
a) the denominator of Eq.~\eqref{eq:infonce} would use all pairs
and not only negative pairs (this variant is a frequently 
used version of InfoNCE), 
b) $1/\tau=\beta$, 
c) the similarity function is the dot product, and
d) $\Bxi$ is used instead of $\Bxi^{\mathrm{new}}$.
Our experiments show that the pattern loss 
can lead to models with comparable performance to 
those trained with cross-entropy loss. The 
according algorithm with pattern loss as alternative
loss is shown in Alg.~\ref{alg:mhn_general}.

We provide a formulation of our 
method as simplified pseudo-code 
in a Python/Pytorch\citep{paszke2019pytorch}-like language (see
Algorithm~\ref{alg:mhn_general}). The pseudo-code provides
a version with two stacked Hopfield layers and 
three possible formulations of the loss function. 

\begin{algorithm}
   \caption{MHN for reaction template prediction (simplified, e.g. skip-connections omitted).}
   \label{alg:mhn_general}
\begin{algorithmic}
   \STATE  {\color{lightblue} \# \texttt{mol\_encoder()}} --- e.g. fully-connected or MPNN. Maps to dimension $d_m$.
   \STATE {\color{lightblue} \# \texttt{template\_encoder1()}}  Maps to dimension $d_{t_1}$.
   \STATE {\color{lightblue} \# \texttt{template\_encoder2()}}  Maps to dimension $d_{t_2}$.
   \STATE {\color{lightblue} \# \texttt{m\_train, t\_train}} --- pair of product molecule and reaction template from training set
   \STATE {\color{lightblue} \# \texttt{T}} --- set of $K$ reaction templates including \texttt{t\_train}
   \STATE {\color{lightblue} \# \texttt{d}} --- dimension of Hopfield space
   \STATE \texttt{\quad}
   \STATE {\color{lightblue} \texttt{\#\# FORWARD PASS}}
   \STATE \texttt{T1\_h = template\_encoder1(T){\color{lightblue} \#[d\_t1,K]}}
   \STATE \texttt{T2\_h = template\_encoder2(T){\color{lightblue} \#[d\_t2,K]}} 
   \STATE \texttt{m\_h = mol\_encoder(m\_train){\color{lightblue} \#[d\_m,1]}} 
   \STATE \texttt{xinew1,\_,\_ = Hopfield(m\_h,T1\_h,dim=d)} 
   \STATE \texttt{xinew,p,X = Hopfield(xinew1,T2\_h,dim=d)}
   \STATE \texttt{p=pool(p,axis=1) {\color{lightblue} \#[K,1]}} 
   \STATE \texttt{\quad}
   \STATE {\color{lightblue} \texttt{\#\# LOSS}}
   \STATE {\color{lightblue} \texttt{\# association loss}}
   \STATE \texttt{label = where(T==t\_train){\color{lightblue} \#[K,1]}} %
   \STATE \texttt{loss = cross\_entropy(p,label)} 
   \STATE {\color{lightblue} \texttt{\# alternative 1: Hopfield loss}}
   \STATE \texttt{L = diag(where(T==t\_train)){\color{lightblue} \#[K,K]}} %
   \STATE \texttt{loss = -log(sum(L@p))}
   \STATE {\color{lightblue} \texttt{\# alternative 2: pattern loss}}
   \STATE \texttt{label = where(T==t\_train){\color{lightblue} \#[K,1]}} %
   \STATE \texttt{pos = X[label] {\color{lightblue} \#[d,1]}}
   \STATE \texttt{neg\_label = where(T!=t\_train){\color{lightblue} \#[K,1]}}
   \STATE \texttt{neg = X[neg\_label] {\color{lightblue} \#[d,K-1]}}
   \STATE \texttt{loss = -InfoNCE(xinew,pos,neg)}
\end{algorithmic}
\end{algorithm}

\end{document}